%% file: main.tex
\documentclass{article}

\usepackage[preprint]{neurips_2026}

\usepackage[utf8]{inputenc} 
\usepackage[T1]{fontenc}    
\usepackage{hyperref}       
\usepackage{url}            
\usepackage{booktabs}       
\usepackage{amsfonts}       
\usepackage{nicefrac}       
\usepackage{microtype}      
\usepackage{xcolor}         

\usepackage{colortbl}
\usepackage{graphicx}       

\usepackage{algorithm}
\usepackage{algpseudocode}

\usepackage{amssymb}
\usepackage{amsmath}
\usepackage{mathrsfs}

\usepackage{multirow}
\usepackage{makecell}
\usepackage{adjustbox}
\usepackage{siunitx}

\usepackage{caption}
\usepackage{subcaption}

\usepackage{pdflscape}

\DeclareMathOperator*{\argmax}{argmax}

\title{Robust Stream Classification using Time Neutralising Decision Trees}

\author{
  Brandon Gower-Winter \\
  Utrecht University \\
  8 Hiedelberglaan, Utrecht, 3584 CS, NL \\
  \texttt{b.gower-winter@uu.nl} \\
  \And 
  Georg Krempl \\
  Utrecht University \\
  8 Hiedelberglaan, Utrecht, 3584 CS, NL \\
  \texttt{g.m.krempl@uu.nl} \\
}

\begin{document}

\maketitle

\begin{abstract}
Data Streams are characterised as a potentially infinite source of data subject to non-stationarity. This non-stationarity, also called Concept Drift, often results in deployed predictive models having to be perpetually retrained as the underlying data generating process of the stream changes over time. Drift Detectors are a class of methods used to identify these changes, triggering these retraining procedures. Historically, Stream Learning has operated under the assumption that instances are only seen once, and cannot be stored for later use. Contemporary work relaxes this assumption, yet drift detection and simple resetting procedures are still ubiquitous.

Building on related work which has shown that traditional Batch Learning algorithms often outperform Stream Learning algorithms when instances can be stored, we introduce Time Neutralising Trees (TNT), a Decision Tree architecture that enables robust stream classification in settings with Concept Drift. During training, TNT will filter out ("neutralise") old training samples that are no longer relevant given the current context of the stream. If those instances become relevant again, TNT will then re-include them. We evaluate TNT, and related algorithms on 24 real and semi-real classification data streams from the USP DS repository. Results show strong evidence that TNT achieves best-in-class performance on a wide-range of data streams with varying concept drift types.
\end{abstract}

\section{Introduction} \label{sect:intro}

In an increasingly digitised world, data streams are at the heart of many online Machine Learning (ML) applications. These streams are often unbounded in size with samples arriving with extreme frequency. These streams also exhibit evolutionary behaviour in that they are often non-stationary. This non-stationarity is known as Concept Drift \cite{GamaZliobaiteBifet2014,KhamassiSayedHammami2018, LuLiuDong2018, ReadZliobaite2025}. In stream prediction settings such as stream classification, concept drift can have severe effects on a deployed model's predictive performance. In some cases, models may need to be perpetually retrained and old instances need to be forgotten if they conflict the current context of the stream. Historically, Change or Drift Detectors are used to identify when a model needs to be reset or updated \cite{LiuLuLiu2018, HinderVaquetHammer2024, LukatsZielinskiHahn2025}. This class of methods is particularly useful in the traditional stream learning setting where instances are only ever seen once and cannot be stored for any period of time.

In recent years, the restriction on storing instances has been criticised \cite{ReadZliobaite2025}, with researchers noting that compute and memory resources are nowhere near as scarce as they used to be when this field was first founded decades ago. In fact, it is almost entirely impossible to effectively use modern ML methods such as Deep Neural Networks or many modern drift detectors \cite{LuLuLiu2025, LukatsZielinskiHahn2025} without relaxing this restriction. Recent work has also shown that in settings where instances can be stored for a period of time, simple retraining or resetting regimes for batch ML algorithms often offers better predictive performance over the duration of a data stream when compared to traditional incremental stream learning methods \cite{GowerGroenKrempl2026}. Furthermore, the use of drift detectors to determine when a model should be reset has been criticised, this is because most drift detectors only tell "When" drift has occurred, but often do not tell "Where" in feature space the drift occurred \cite{LuLiuDong2018}.

In this work, we aim to demonstrate that the somewhat arbitrary sample storage restriction, and over-reliance on drift detectors as reset mechanisms limits one's ability to develop robust stream classification algorithms. We do this by introducing a new method called the Time Neutralising Tree (TNT). In short, at every split decision node during training, TNT will first look to see if splitting on time (the arrival time indices of the instances) is the most informative split decision. If so, TNT will filter out (neutralise) older instances that do not match the label distributions of recent instances. In doing so, TNT is able to contextually determine which instances are relevant at any given point in the data stream, allowing it to adapt to concept drift phenomena without an explicit drift detection method. The contributions of our paper are three-fold:

\begin{enumerate}
    \item We refine the definition of concept drift by conditioning its occurrence on sub-regions of the feature space, rather than the global perspective taken in most literature. Through identifying regions in the feature space that have drifted, we are able to provide formal definitions, using Information Theory, for when retraining or resetting are appropriate actions to take.
    
    \item Using these theoretical findings, we introduce Time Neutralising Trees (TNT) which dynamically filter out instances that are irrelevant to learn on given the current context of the stream.

    \item We evaluate TNT and other relevant methods on 24 real and semi-real data stream classification tasks taken from the USP DS repository \cite{SouzaReisMaletzke2020}. Results show that TNT performs better than most other methods. We hope this inspires further investigation into stream learning methods that move away from the assumption that data storage is not allowed in stream learning. To the best of our knowledge, this paper provides the the most comprehensive evaluation of decision tree algorithms for data stream classification tasks. 
\end{enumerate}

The paper proceeds as follows: Section \ref{sect:background} provides a summary of related work, Section \ref{sect:methodology} describes TNT, Section \ref{sect:exp_design} outlines our experimental design, Section \ref{sect:results} reports the results of our experiments and the paper concludes with a discussion and outline of future work (Section \ref{sect:conclusions}). All source code for this work is made available at: \url{https://github.com/BrandonGower-Winter/TNT}.

\section{Background and Related Work} \label{sect:background}

A data stream may be defined as a mode of access to a potentially infinite sequence of instances generated by some concept and delivered to a learning algorithm by an Instance-Delivery-Process (IDP) \cite{ReadZliobaite2025}. In a traditional  supervised machine learning task, it is assumed that the underlying data generating process remains constant throughout time $(x_0,y_0), (x_1,y_2), \ldots (x_n,y_n) \sim P(X, Y)$ \cite{HinderVaquetHammer2024}, where $x_i$ and $y_i$ are the i-th instance's feature and label, sampled from stationary concept $P(X,Y)$.

In stream learning, the underlying data generating process may change over time such that each instance is potentially sampled from a different time-dependent concept $(x_0,y_0), (x_1,y_2), \ldots (x_i,y_i) \ldots \sim P_t(X, Y)$ with  timestamp (or instance index) $t$. Formally, concept drift is then defined as a change in concept from one timestep to another $P_t(X,Y) \neq P_{t+1}(X,Y)$, although some have argued that it would be more appropriate to model concept drift as a distribution process \cite{HinderVaquetHammer2024}. Concept drift is said to typically occur in either the posterior distribution $P(Y|X)$ or in the covariates $P(X)$ with posterior drift typically requiring retraining or resetting while the covariate shift is only correlated with retraining or resetting \cite{ReadZliobaite2025}. There also exist several patterns that describe concept drift. For example, drift occurring over a short period of time is called abrupt drift. Drift occurring over a longer period of time is called incremental or gradual drift (the distinction between the two is not relevant for this work, see Lu et al. \cite{LuLiuDong2018}). Drift occurring in a cyclical pattern (e.g., public transportation usage on weekdays vs. weekends), is known as recurring drift.

\subsection{Drift Detectors}

Formally, a drift detector can be defined as a binary classification task, where given a set of descriptors (descriptions of the observed data), a drift detector outputs a binary "Yes"/"No" depending on whether drift has been detected or not: $\mathcal{M}: W \mapsto \{0, 1\}$ where $\mathcal{M}$ is the drift detector and $W$ is a window that contains the descriptions of the data. Drift detection typically consists of 4 stages \cite{LuLiuDong2018, HinderVaquetHammer2024}. Stage 1 selects the instances to send to the drift detector (window-based methods are most common). Stage 2 uses the selected instances to build descriptors of the data. Stage 3 computes the dissimilarity of the descriptors, and Stage 4 calibrates the dissimilarity which identifies drift in the selected instances.

Drift detectors are typically categorised as either supervised or unsupervised. Supervised detectors (e.g., DDM \cite{GamaMedasCastillo2004}, ADWIN \cite{BifetGavalda2007}, STUDD \cite{CerqueiraGomesBifet2023}, PUDD \cite{LuLuLiu2025}) typically use class labels and the predictions made by a deployed model to detect drift while unsupervised detectors (e.g., D3 \cite{GozuaccikBuyukccakirBonab2019}, IBDD \cite{SouzaParmezanChowdhury2021}) only require access to instance features. Both categories of detectors are limited in the types of drift they can detect \cite{Zliobaite2010} with supervised detectors only reliable when labelled data is present and if changes in the performance of the deployed predictor reliably indicate drift \cite{KunchevaZliobaite2009,HinderVaquetBrinkrolf2023, ReadZliobaite2025}. 
Conversely, unsupervised detectors can only detect changes in the covariates $P(X)$ and not in the posterior distribution $P(Y|X)$. Recently, drift detection has received criticism for being limited in the information they provide, as well as being subject to the Window Dilemma \cite{GowerGroenKrempl2026}. Most detectors only tell you "When" drift has occurred, but not "Where" in the feature space it has occurred \cite{LuLiuDong2018}. As we will show, feature space conditioning is imperative in settings with concept drift, as it informs a deployed classifier on how it should adapt.

\subsection{Incremental Decision Trees}

Incremental Decision Trees and their derivates are the most commonly-used methods in stream learning. The most famous of which is the Hoeffding Tree (HT) \cite{HultenSpencerDomingos2001} which uses the Hoeffding Bound to determine tree splits. Since its inception newer incremental decision tree such as the Extremely Fast Decision Tree (EFDT) \cite{ManapragadaWebbSalehi2018} and PLASTIC \cite{HeydenGomesFouche2024} were developed which are less conservative when making split decisions. Even state-of-the-art stream methods (mostly ensembles) such as Adaptive Random Forests \cite{GomesBifetRead2017}, Aggregated Mondrian Forests \cite{MourtadaGaiffasScornet2021}, Streaming Random Patches \cite{GomesReadBifet2019} and Streaming Gradient Boosted Trees \cite{GunasekaraPfahringerGomes2024} use incremental decision trees as their base models. Given that incremental decision trees are not drift-aware, they are typically combined with a drift detector (commonly DDM \cite{GamaMedasCastillo2004} or ADWIN \cite{BifetGavalda2007}). A drift-aware incremental decision tree will wait for a drift-\emph{warning} signal from its drift detector to start training new replacement tree in the background, and upon a drift-\emph{reset} signal will replace the old with the background tree. This procedure represents a global reset mechanism by which the entire tree is replaced. There are tree variants such as the Hoeffding Adaptive Tree Drift (HAT) \cite{BifetGalvalda2009} and Locally Adaptive Streaming Tree (LAST) \cite{AssisBarddalEnembreck2024} which place drift detectors nodes in the tree structure. HAT uses a detector to reset subtrees when a drift-signal is received while LAST uses drift signals to determine when to split. These methods represent a more conservative approach where only drifted regions of the tree are replaced. 

In the standard stream learning setting, when a drift event replaces or resets some number of components in a learning algorithm, the learned information contained in those components is typically lost forever. This is clearly undesirable in settings with recurring contexts where forgotten information may once again be useful for making accurate predictions. Typical solutions to this recurring contexts problem utilise archival techniques such as those FLORA \cite{WidmerKubat1996}, and COPL \cite{Koychev2002}. In literature however, decision tree architectures rarely explore archival strategies.

Nevertheless, it has been argued that much of the success of these methods is due to the Naive Bayes models that occupy the leaf nodes of these decision trees, rather that the tree algorithms themselves. This is because incremental decision trees are typically quite conservative when it comes to making decision splits \cite{ReadZliobaite2025}. Furthermore, recent work has shown that in settings where instances can be stored for some period of time, even simple retraining or resetting regimes using batch decision trees can outperform their stream learning counterparts \cite{GowerGroenKrempl2026}. In this work, we take this one step further by showing that in settings where instances can be stored for some period of time, decision trees architectures can explicitly determine which instances are relevant to train on given the current context of the data stream, thus enabling them to be drift adaptive.

\section{Methodology} \label{sect:methodology}

In this section, we introduce Time Neutralising Trees (TNT), a concept drift-robust batch learning method for stream settings where instances can be stored for period of time. Section \ref{sect:method_motive} introduces the theory behind TNT, and Section \ref{sect:method_TNT} describes the method.

\subsection{Motivation} \label{sect:method_motive}
\begin{figure}[ht]
  \centering
  \includegraphics[width=\linewidth]{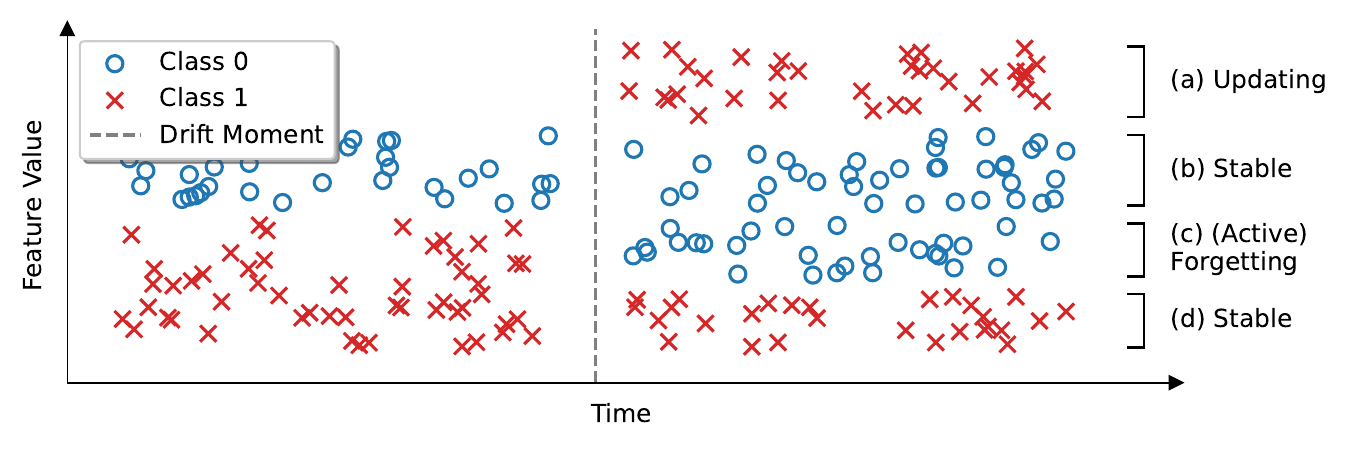}
  \caption{illustrates why feature subspaces must be taken into account when learning in settings with concept drift. Given a univariate stream with a single drift moment (dashed line) and assuming a model was trained on data prior to the drift, different actions need to be taken to update the model on the post-drift data, depending on the location in the feature space. Subspace (a) is entirely new, so the optimal action is to learn in that new region. Subspaces (b) and (d) are stable (i.e., no drift), so no action needs to be taken (the model still classifies instances in those regions correctly). Subspace (c) however, requires that the model actively forgets instances from the past before learning on instances from the current time context (i.e., the pre-drift context conflicts with the post-drift context).}
  \label{fig:TNT_motive}
\end{figure}

We first discuss the theory behind TNT. Figure \ref{fig:TNT_motive} illustrates a synthetic univariate stream with a concept drift at time some arbitrary time point $t$. Given a model $\theta$ trained on instances prior to the drift event, we want to decide how $\theta$ should be updated to adapt to the drift. As shown, updating $\theta$ depends on where in the feature space we observe drift. In subspaces (b) and (d), no drift has occurred and $\theta$ does not need to be updated. Subspace (a) is entirely new (i.e., there are no past instances in this subspace) so the optimal action is to simply update $\theta$ on those new instances. However, subspace (c) represents a region where past instances are no longer relevant in the current context of the stream (i.e., they degrade prediction quality). Here the optimal action is to actively forget those old instances and update $\theta$ on the new data. When viewed holistically, subspaces (a, b, d) require no further action beyond retraining, which given a sufficient set of instances $X$ and labels $Y$, any suitable batch learner can adapt to. 
We will now focus on scenarios like subspace (c) where active forgetting is useful.

Consider a non-stationary data stream for a supervised learning task $S = \{(x_0,y_0), (x_1,y_1), \ldots, (x_n,y_n), \ldots\}$ where $x_i \in \mathcal{X}$ and $y_i \in \mathcal{Y}$. In non-stationary streams, an instance $(x_i, y_i) \sim \mathcal{D}_{t=i}$ where $\mathcal{D}_t$ is a distribution map of time to a concept $\mathcal{D} : t \to P_t(X,Y)$. If a drift occurs at timestep $t_{drift}$, $\mathcal{D}_{t < t_{drift}} \neq \mathcal{D}_{t \geq t_{drift}}$.

In a stationary stream, expected Mutual Information (MI) $I(X;Y) = H(Y) - H(Y|X)$ remains constant throughout the lifetime of a stream. In a non-stationary stream, the MI of the entire stream is dependent on the global distribution which is a weighted mixture of the $K = |\mathcal{D}|$ \emph{distinct} concepts that have occurred in the stream's lifetime:

\begin{equation}
    \bar{P}(X, Y) = \sum_{k=1}^K w_k P_k(X, Y)
\end{equation}

where $w_k$ is the normalised duration of concept $P_k(X,Y)$. If the number of distinct concepts $K \to \infty$, then the MI of the stream tends to 0 (see Appendix \ref{sect:app_MI_EMP}):

\begin{equation}\label{eq:MI_limit}
    \lim_{K \to \infty} I_{\bar{P}}(X;Y) = 0
\end{equation}

Note that we are particular in our usage of the word \emph{distinct}. Two streams constructed using concepts $A \to B \to C$ and $A \to B \to A \to C$ resp. have the same number of \emph{distinct} concepts. We have included empirical simulations that support Eq. \ref{eq:MI_limit} in the Appendix (\ref{sect:app_MI_EMP}). Given that MI is convex with respect to the conditionals $P(Y|X)$ and fixed distribution $P(X)$ \cite{CoverThomas2006}, Jensen's Inequality states that the expected MI of the entire stream must be less than or equal to the expected MI of the \emph{distinct} concepts in that stream:

\begin{equation} \label{eq:jensens}
    I_{\bar{P}}(X;Y) \leq \sum_{k=1}^K w_kI_{k}(X;Y)
\end{equation}

Using Eq. \ref{eq:jensens}, we now have a method for solving the active forgetting problem in subspace of Fig \ref{fig:TNT_motive}. Given a set of timesteps $T = \{0, 1,\ldots, n, \ldots\}$, our goal is find some time interval $\mathcal{T}^* \subseteq T$ where:

\begin{equation}
    \mathcal{T}^* = \argmax_{\mathcal{T} \in \mathcal{P}(T)} \; I(X;Y|\mathcal{T}) - I_{\bar{P}}(X;Y)
\end{equation}

In practice, we would want $\mathcal{T}^*$ to include the most recent timestep $n \in \mathcal{T}^*$. That way we ensure the time interval we find is relevant to the current stream context. Note that, fundamentally, supervised drift detectors already approximate this process, but without the ability to recover instances once they are forgotten due to the inability of most stream learning algorithms to store past data. In a modern setting where we do have the ability to store data, we can approximate $\mathcal{T}^*$ more accurately. Decision Trees are naturally equipped to do this as they maximize information gain through partitioning the feature space. This is the primary motivation for TNT, by treating time $T$ as an additional feature, we are able to find subspaces where active forgetting is useful, thus allowing for a drift robust batch learner to be trained without an explicit drift detection mechanism.

\subsection{Time Neutralising Trees} \label{sect:method_TNT}

In this Section, we describe the Time Neutralising Tree (TNT).
Like traditional decision trees, given a set of instances $X$, and labels $Y$, TNT partitions the feature space by impurity reduction (TNT uses Information Gain, see Alg. \ref{alg:tnt_build}). This process continues until some stopping condition is reached for all possible decision paths (TNT uses minimum samples $N_{min}$). 
Rather than maintaining an incremental decision tree, TNT aims to create a new decision tree optimised to the current time point. To this end, from all currently available data TNT filters out (neutralises) the irrelevant data during the training process. 
Thus, unlike traditional decision trees, \emph{during training} TNT also accepts the time indices of the instances $T$. These could be the actual clock times of each instance's arrival or a simple index denoting the instance's position in the stream. At every decision node, TNT will also calculate the best possible split on $T$. Then, using the following rule:

\begin{equation}
    \text{BestTimeSplit}(T, Y) > \text{BestFeatureSplit}(X, Y) + \lambda
\end{equation}

TNT determines if time is a more informative feature than the best feature split. If that is the case, TNT begins the filtering process. $\lambda \in[0, 1)$ controls the reactivity of TNT. If $\lambda$ is close to 0.0, it will filter out instances as soon as time is the most relevant feature. As $\lambda$ increases, TNT becomes more reluctant to filter historical instances unless the IG from such a split is sufficiently large. We include $\lambda$ as a tuneable parameter because streams can exhibit various phenomena such as label autocorrelation and temporal dependencies \cite{SouzaReisMaletzke2020} which necessitate different degrees of reactivity.

\begin{figure}[ht]
     \centering
     \begin{subfigure}{\textwidth}
         \begin{subfigure}{0.69\textwidth}
             \centering
             \includegraphics[width=\textwidth]{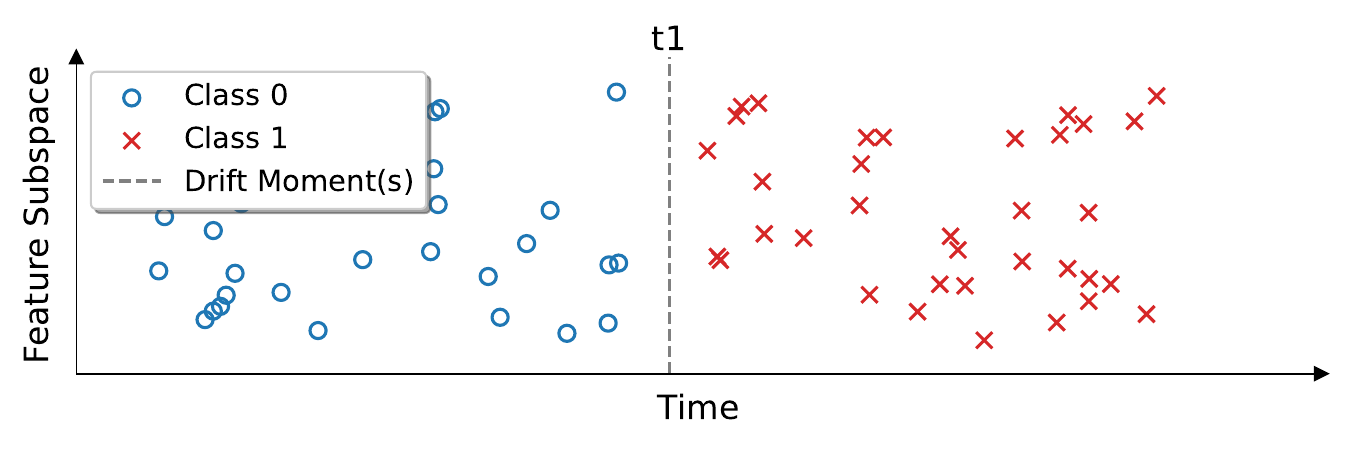}
             \label{fig:time_filter_stream_one}
         \end{subfigure}
         \hfill
         \begin{subfigure}[b]{0.29\textwidth}
             \centering
             \includegraphics[width=\textwidth]{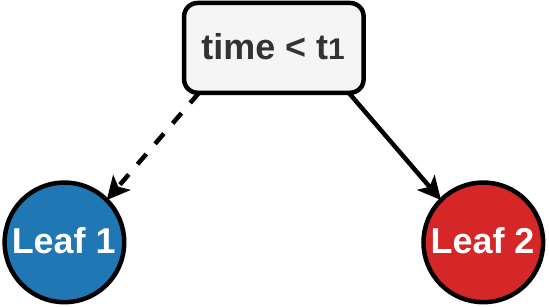}
            \label{fig:time_filter_tree_one}
        \end{subfigure}
        \caption{}
        \label{fig:time_filter_one}
     \end{subfigure}
     \begin{subfigure}{\textwidth}
         \begin{subfigure}{0.69\textwidth}
             \centering
             \includegraphics[width=\textwidth]{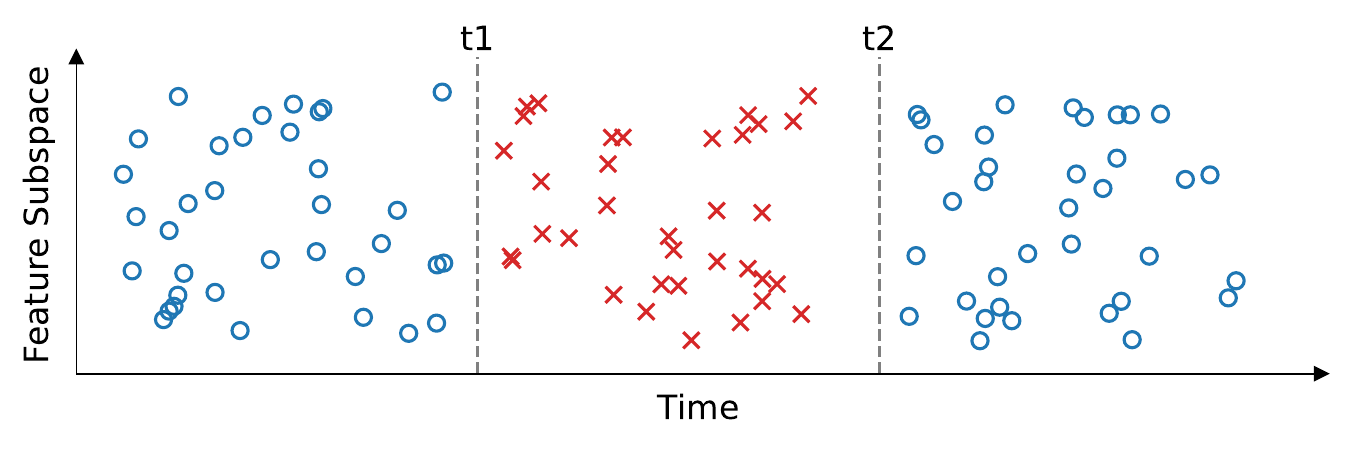}
             \label{fig:time_filter_stream_two}
         \end{subfigure}
         \hfill
         \begin{subfigure}{0.29\textwidth}
             \centering
             \includegraphics[width=\textwidth]{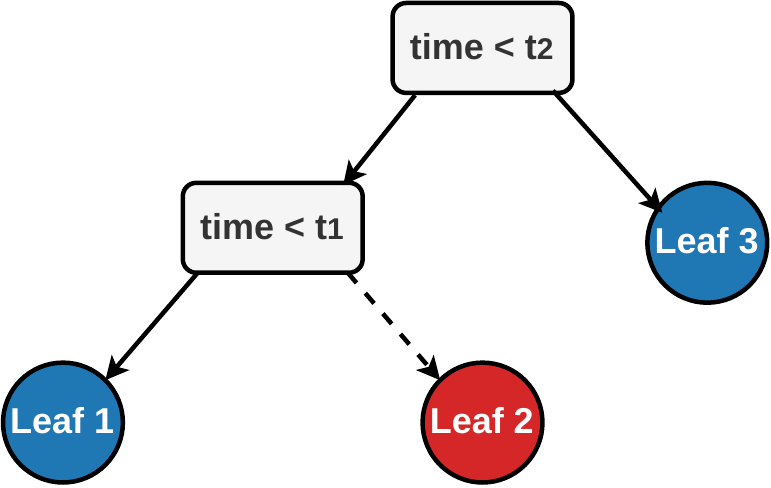}
             \label{fig:time_filter_tree_two}
         \end{subfigure}
         \caption{}
         \label{fig:time_filter_two}
     \end{subfigure}
     \caption{illustrates the time filtering process used by TNT. Here we see an arbitrary subspace of a univariate stream at two different moments in time. Figure (a) shows the stream after one drift has occurred at time $t1$. The corresponding time filtering tree is shown on the right whereby the old instances (class 0, blue) are filtered out (dashed line) and no longer considered by TNT when determining additional split nodes during training. Figure (b) shows the same subspace at a later moment in time where a reoccurring drift is seen at $t2$. The corresponding time filtering tree now filters out the previously selected instances (class 1, red, dashed line) and keeps the class 0 instances for further training.}
     \label{fig:time_filtering}
\end{figure}

When the filtering process is initiated on a decision node (Alg. \ref{alg:tnt_temporal_filter}), TNT will create a subtree that splits only on time until the node purity, minimum samples ($N_{min}$) or some minimum level of information gain $\tau$ are reached. Then, using the right most leaf node (i.e., node that most accurately represents the current stream context for that subspace) of the time tree as reference, a Mann-Whitney U test is performed ($p=\alpha_{mwu}$) on each other leaf node's label distribution. We use the Mann-Whitney U test because it carries the fewest assumptions about the label distribution, but it could be replaced with another relevant statistical test. If the U test fails (i.e., $p > \alpha_{mwu}$), that leaf node is merged with the reference leaf node as it contains data that is relevant in the current stream context (i.e., a node with a label distribution similar to the reference node's label distribution contains information that is relevant in the current stream context). All leaf nodes in the time subtree that are not merged are discarded, and all merged nodes are used to create a set of filtered instances that are returned to the main tree (See Fig. \ref{fig:time_filtering} for visual aid.). Note: we cannot only perform a single split on time because relevant instances may exist in disjoint subsets of time (e.g. reoccurring concepts as in Fig. \ref{fig:time_filter_two}), so we must continually split during the filtering process until the stopping conditions are met. It is this filtering process that allows TNT to "neutralise" the effects of concept drift in areas where active forgetting is useful.

Once a filtering step is performed, the filtered instances, their labels, and time indices are returned (lines 13-14 in Alg. \ref{alg:tnt_build}). TNT will then recalculate the best feature split for the filtered instances and continue the training process. Note: the time indices $T$ are only used during the training process, not during inference. That is because any splits on time immediately become irrelevant as new instances arrive at inference time. Time is only used to filter out training instances.

\section{Experimental Design} \label{sect:exp_design}

For our evaluation, we make use of 24 real and semi-real data streams taken from the USP DS repository \cite{SouzaReisMaletzke2020}. The streams are binary and multi-class classification tasks that cover a wide-range of application domains such as Urban Development, Medicine, and Epidemiology. These stream vary in the number of features they possess (between 2 and 3600) and sample size (between 503 and 581,012).
A summary of these streams are included in the Appendix (\ref{sect:app_exp_details}). 

As is standard practice in stream learning, we make use of the test-then-train paradigm for evaluation. Thereby, deployed models first predict on instances before they train on them. For baselines, we make use of Majority Class (MC) and Last Class (LC) classifiers. These are also sometimes called Prior and NoChange classifiers which predict the majority and last label seen, resp. These two baselines reveal class imbalance and autocorrelation in the labels which are common in data streams. We evaluate the following stream learning algorithms: Hoeffding Tree (HT) \cite{HultenSpencerDomingos2001}, Extremely Fast Decision Tree (EFDT) \cite{ManapragadaWebbSalehi2018}, PLASTIC (PLS) \cite{HeydenGomesFouche2024}, Hoeffding Adaptive Tree (HAT) \cite{BifetGalvalda2009} and Locally Adaptive Streaming Tree (LAST) \cite{AssisBarddalEnembreck2024}. HAT and LAST are the only drift-aware models. Given this, we also make use of two supervised drift detectors: ADWIN \cite{BifetGavalda2007} and DDM \cite{GamaMedasCastillo2004} and one unsupervised drift detector: D3 \cite{GozuaccikBuyukccakirBonab2019}. All three detectors are used to reset a Hoeffding Tree when they detect drift. In our notation, $\text{HT}_{\text{DD}}$ thus refers to a Hoeffding Tree that is reset completely when the drift detector (DD, which is either ADWIN, DDM or D3) signals that drift has been detected. We do not evaluate more unsupervised detectors because their comparison is somewhat unfair. Unsupervised detectors typically require that you predefine their detection behaviour. Here we use D3 to reset a model, but unsupervised detectors can also be used as monitoring algorithms to determine when new instance labels should be acquired, which is beyond the scope of this work. For all stream learning methods, once they make a prediction on a previously unseen instance, they immediately train on it in accordance with test-then-train procedure.

For the batch learners, we make use of standard batch Decision Trees (DT) and sliding windows. Given that we cannot train batch DTs incrementally, they are retrained from scratch every 50 instances. The DTs we use are: DT$_{50}$, DT$_{100}$, DT$_{250}$, DT$_{500}$, and DT$_{\infty}$. That is, a decision tree DT$_n$ is trained on the last $n$ instances of the stream. This gives us an broad overview of the types of performance one can achieve when data can stored a various sizes. For DT$_{\infty}$, it is trained on all previously seen instances in the stream. For TNT, it follows the same training procedure as DT$_{\infty}$. However, we acknowledge that storing all data over an entire stream is infeasible in certain settings so we also evaluate TNT$_{500}$ which follows the same training procedure as DT$_{500}$ (i.e., every 50 instances, the model is retrained on the last 500 instances). With this selection, we believe our evaluation covers a broad range of learning strategies over varying time scales and data storage demands.

During evaluation, we track each model's Cohen's Kappa Score \cite{Cohen1960}, recording it every 50 instances over the last 50 instances. The exception to this are the first 50 instances for which no metrics are tracked as this period is used as a warm start for all methods. We use kappa score because accuracy is not always an effective metric in stream settings  \cite{GomesBarddalEnembreck2017}. For all stochastic methods, 10 replicates per stream are run using a pseudo-random number generator and provided seeds to ensure reproducibility. Fractional ranking is used to determine the ordering of methods per stream (i.e., if the kappa for some methods are [85.0, 80.0, 80.0, 60.0] their corresponding ranks are [1, 2.5, 2.5, 4]). Each method is ranked by mean kappa score, and a Mann-Whitney U test ($p=0.05$) with Holm-Bonferroni correction is used to determine if two methods are statistically different. If they are not, they are given the same rank. Once ranking is complete, a critical difference diagram ($p=0.05$) is constructed \cite{Demsar2006, BenavoliCoraniMangili2016} to determine statistical significant rankings across the evaluated methods. Additional experimental setup details are included in the Appendix (\ref{sect:app_exp_details}).

\subsection{Hyper-parameter Tuning and Sensitivity Analysis} \label{sect:exp_design_param_tune}

Before evaluating TNT, we sought to understand how its 5 tunable parameters affect its predictive performance. They are: filter factor $\lambda$, minimum split node samples $N_{min}$, minimum time node entropy split $\tau$, confidence threshold for determining time node merges $p_{mwu}$, and the maximum number of threshold splits to consider per feature $\text{thresholds}_{max}$. To avoid data leakage and bias \cite{SouzaReisMaletzke2020}, we parameter tune TNT on 5 synthetic streams which cover a variety of drift types. They are: \verb|Hyperplane| \cite{HultenSpencerDomingos2001}, \verb|RandomRBFDrift| \cite{Montiel2021}, \verb|SEA| \cite{StreetKim2001}, \verb|STAGGER| \cite{SchlimmerGranger1986} and a custom stream \verb|RotatingMoons| (\ref{sect:app_rot_moons}). After some initial exploration of the hyper-parameters, we used Quasi-Monte Carlo Latin Hypercube Sampling \cite{Niederreiter1992} to generate $100$ parameter configurations which we explored across all synthetic streams, recording the final accuracy of the model per configuration per stream. For each stream, we then train a Random Forest \cite{Breiman2001} regressor using the parameter configurations as input and the final accuracy as a target. From each trained Forest, we extract feature importances (\ref{sect:app_param_tune_fi}). We also plot each parameter and obtained accuracy per stream using LOWESS regression to analyse per feature level dependencies (\ref{sect:app_param_tune_sa}).

Results indicate that $N_{min}$ and $\text{thresholds}_{max}$ are generally the most important hyper-parameters across the synthetic streams. This is understandable as they both control the model's ability to generalise. For \verb|Hyperplane|, a low filter factor $\lambda$ is the most significant parameter. This is because active forgetting is pivotal in this stream where old instances rapidly become outdated. For the other streams, there is a evidence to suggest that too low of a filter factor (which increases instance filtering) is detrimental to the TNT's performance, confirming our intuition that there needs some balance between reactivity and preservation in the time filtering process. Interestingly, $\lambda$ was not as pivotal in determining model performance as we expected. We discuss this further in the postmortem (\ref{sect:app_postmortem_param_tune}). 

For the final step, we performed an ablation on TNT whereby we replaced impure leaf nodes with either a Logistic Regression (LR) or Naive Bayes (NB) model as is done in incremental decision trees. This is in contrast to the majority class prediction strategy used by most batch decision trees. Results (\ref{sect:app_param_tune_la}) indicated that using LR on non-pure leaf nodes gave a slight performance improvement as it mitigates the effects of poorly chosen split node minimums $N_{min}$ and $\text{thresholds}_{max}$ which, as noted, greatly affect TNT's generalisability. A complete description of the parameter tuning process is included in Appendix (\ref{sect:app_param_tune}), and Table \ref{tab:hyperparameters} lists the final hyper-parameters used in our evaluation.

\begin{table}
  \caption{Hyper-parameters and their respective values used in our evaluation.}
  \label{tab:hyperparameters}
  \centering
  \begin{tabular}{l|l|l}
    \toprule
    Name     & Description     & Value \\
    \midrule
    \rowcolor[gray]{0.9}
    $\lambda$ & Controls the time filtering sensitivity.  & 0.20 \\
    $N_{min}$ & Minimum number of samples needed to make a spit decision. & 20 \\
    \rowcolor[gray]{0.9}
    $\tau$ & Stopping threshold for when time splits are not informative. & 0.05 \\
    $\alpha_{mwu}$ & Threshold used to merge time leaf nodes using Mann-Whitney U Test. & 0.05 \\
    \rowcolor[gray]{0.9}
    thresholds$_{max}$ & Number of split candidates generated per feature. & 15 \\
    leaf model & Predictor that replaces non-pure leaf nodes. & LR \\
    \bottomrule
  \end{tabular}
\end{table}

\begin{figure}[ht]
  \centering
  \includegraphics[width=\linewidth]{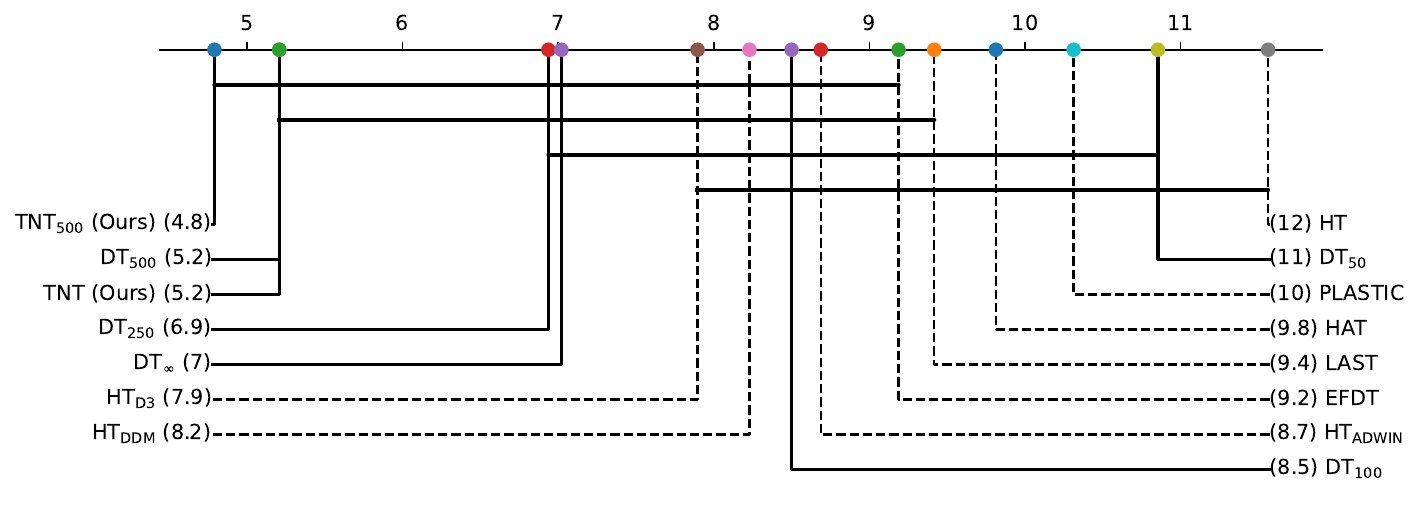}
  \caption{Critical Difference Diagram ($p=0.05$) of all methods across all streams investigated. A lower rank is better. Dashed lines represent Stream Learning methods while solid lines represent Batch Learning methods. Solid horizontal bars represent cliques for which no statistical significance could be found. Baselines MC and LC are omitted for clarity.}
  \label{fig:cd_diagram}
\end{figure}

\section{Results} \label{sect:results}

Table \ref{tab:results_main} reports on the average Kappa score for all models across all streams investigated, and Fig. \ref{fig:cd_diagram} summarises the results with a CD diagram. Overall, results show a clear separation in the performance between batch learning DTs and the incremental DTs. All incremental DTs rank noticeably lower than the best performing batch DTs (TNT$_{500}$, TNT and DT$_{500}$ resp.). These results further echo the findings in Gower-Winter et al. \cite{GowerGroenKrempl2026}, if training instances can be stored for any period of time, it is often better to simply retrain a model on those instances than use an incremental DT. Our results indicate as few as $250$ samples is enough when evaluating on the USP DS repository.

\begin{figure}[ht]
     \centering
     \begin{subfigure}{\textwidth}
         \centering
         \includegraphics[width=\textwidth]{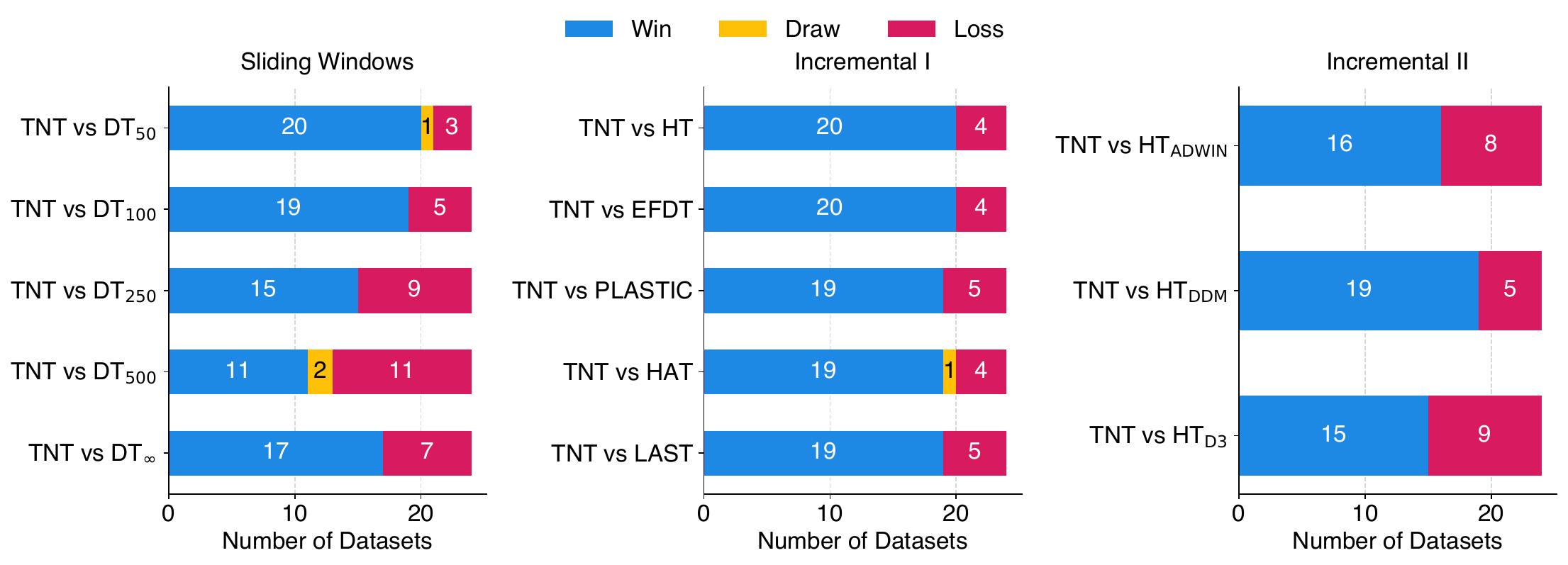}
         \label{fig:wl_TNT}
         \caption{}
     \end{subfigure}
     \hfill
     \begin{subfigure}[b]{\textwidth}
         \centering
         \includegraphics[width=\textwidth]{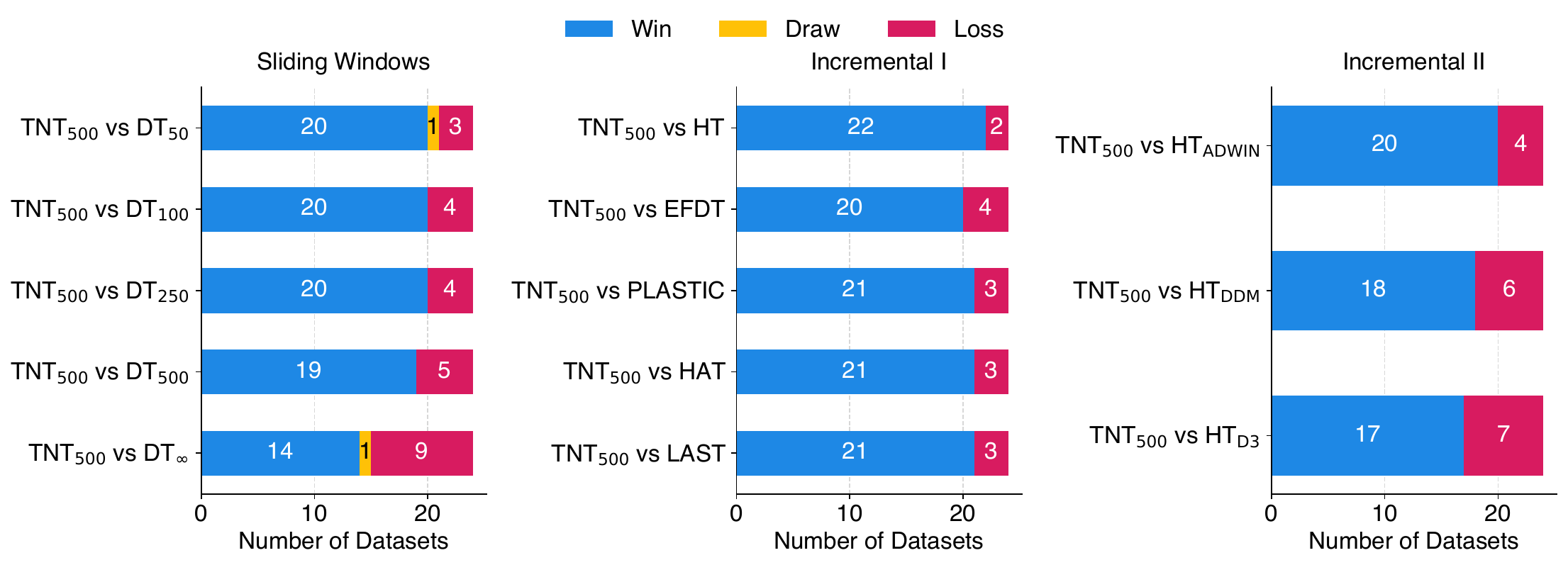}
        \label{fig:wl_TNT500}
        \caption{}
    \end{subfigure}
     \caption{the win-loss ratio of TNT (a) and TNT$_{500}$ (b) on the USP DS repository benchmark against the other methods investigated in this work. A win means that the method on the left had a significantly greater Kappa score the method on the right. The first column reports on the batch (sliding window) learners and columns two and three report on the incremental learners.}
     \label{fig:wl_ratios}
\end{figure}

TNT performed better than expected with both variants (TNT$_{500}$, TNT) achieving mean ranks (4.8, 5.2 resp.) higher than their traditional batch DT equivalents (DT$_{500}$, DT$_{\infty}$, ranks 5.2, 7 resp.). This shows that TNT adapts to the current context of the stream effectively (\ref{sect:app_extra_results}) . This is particularly noticeable on the \verb|GasSensorArray| (GS), \verb|StartLightCurves| (SL) and \verb|Yoga| (YG) streams. TNT$_{500}$ on average ranked higher than TNT implying that passive forgetting may still be useful even when using active forgetting strategies like the instance filtering used by TNT. This is further supported by the higher rank of DT$_{500}$ relative to DT$_\infty$. One possible reason for this is that fewer, but still sufficient, samples prevent the batch DTs from overfitting on historical data. For TNT, passive forgetting using a sliding window also counteracts some loss in predictive performance that can occur when $\lambda$ is tuned too conservatively. Regardless, Fig. \ref{fig:wl_ratios} shows that both variants of TNT explored in this work outperform both the traditional batch and incremental decision trees in the vast majority of cases.

For the incremental DTs, there are several streams where they noticeably outperform the batch DTs. They are Electricity (EL), Forest Cover Type (FC), and Outdoor Objects (OD). This is discussed in more detail in the postmortem (\ref{sect:app_postmortem_incremental_learners}), but these also happen to be the data streams with high autocorrelation in the labels and/or streams that have been extensively studied throughout the history of stream learning. For every stream that the incremental learners excelled at, there are equally as many streams with high feature counts or class imbalances for which they are unable to generalise at all (e.g., Cancer (CA), Heartbeats (HB), and Ozone (OZ)). The results for the incremental DTs are also inflated because they are able to train immediately after making a prediction, ensuring they are always as up-to-date. In contrast, the Batch DTs must make 50 predictions before they are allowed to update themselves. Related work \cite{GowerGroenKrempl2026} has shown that the predictive performance of incremental learners decreases (sometimes significantly) when they are subject to verification latency \cite{KremplZliobaiteBrzezinski2014}. Nevertheless, our results are clear. When historical instances can be stored for a period of time, batch learning methods typically outperform their incremental learning counterparts. With this new framing, TNT stands as a best-in-class, drift-robust stream classifier.

\subsection{TNT as a Preprocessing Step} \label{sect:result_TNT_preprocessing}

The results from our primary investigation indicate that TNT performs best-in-class for single decision trees. However, state-of-the-art ensembles, such as Adaptive Random Forests \cite{GomesBifetRead2017} and Streaming Random Patches \cite{GomesReadBifet2019} out perform single decision trees in most, memory unconstrained, settings. During the development of TNT we found that simply creating an ensemble of TNT trees was not an effective strategy as it results in many repeated computations across the TNT base models. Thus, we explicitly leave the development of a TNT ensemble for future work. Nevertheless, this section demonstrates that TNT can be used as a preprocessing step to enable more complex batch learning architectures such as Random Forests \cite{Breiman2001} and Gradient Boosting models \cite{Friedman2001, Friedman2002} to be drift-adaptive.

\begin{table}
  \caption{Average Kappa Scores of batch learning algorithms (left) with and without using TNT as a preprocessing step to filter out irrelevant samples before training, compared to state-of-the-art stream learning ensembles (right). Kappa scores are averaged over 10 replicates per method on the Hyperplane data stream.}
  \label{tab:TNT_preprocessing}
  \centering
  \begin{tabular}{l|c|c}
    \toprule
    \textbf{Method} & \textbf{Kappa without TNT} & \textbf{Kappa with TNT} \\
    \midrule
    \rowcolor[gray]{0.9}
    \textbf{TNT} & 0.72.  & - \\
    \textbf{DT} & 0.38 & 0.73 \\
    \rowcolor[gray]{0.9}
    \textbf{RF} & 0.39 & 0.76 \\
    \textbf{GB} & 0.38 & 0.75 \\
    \rowcolor[gray]{0.9}
    \textbf{MLP} & 0.39 & 0.74 \\
    \textbf{DNN} & 0.39 & 0.75 \\
    \bottomrule
  \end{tabular}
  \quad
  \begin{tabular}{l|c}
    \toprule
    \textbf{Method}     & \textbf{Kappa} \\
    \midrule
    \rowcolor[gray]{0.9}
    \textbf{AMF}	& 0.52 \\
    \textbf{ARF}	& 0.72 \\
    \rowcolor[gray]{0.9}
    \textbf{SRP}	& 0.76 \\
    \bottomrule
  \end{tabular}
\end{table}

Table \ref{tab:TNT_preprocessing} reports the results of comparing traditional Decision Tree (DT), Random Forest (RF), Gradient Boosting (GB), Multi-Layer Perceptron (MLP), and Deep Neural Network (DNN) architectures with and without the use of TNT as a preprocessing step. When TNT is used for preprocessing, all samples that TNT keeps (i.e., does not filter away) are used to train the primary architecture. We also compare the results of this preprocessing to state-of-the-art streaming ensembles ARF, SRP and Aggregated Mondrian Forests (AMF) \cite{MourtadaGaiffasScornet2021}. We use the \verb|Hyperplane| stream evaluated under the same conditions as the main experiments. We use \verb|Hyperplane| because it simulates pure posterior drift (change in P(Y|X)), enabling us to isolate the effectiveness of the TNT preprocessing step at filtering away irrelevant training samples. Results show that TNT enables models, with no drift adaptation capabilities to match the match the performance of state-of-the-art stream learning algorithms.

\section{Conclusions and Future Work} \label{sect:conclusions}

This paper introduces Time Neutralising Trees (TNT), a batch learning decision tree architecture that is robust to concept drift. TNT uses the time indices of instances as a feature during training, whereby training samples that are no longer relevant to the current context of the stream are filtered ("neutralised") out. We evaluated TNT on 24 real and semi-real data streams from the USP DS repository \cite{SouzaReisMaletzke2020} which showed that TNT achieves best-in-class performance when compared to other decision tree architectures. The main takeaway from this paper is that when historical data can be kept for some period of time, batch learning strategies outperform their stream learning counterparts. In such settings, TNT stands as viable candidate for creating drift robust stream classifiers.

Future work will address the this work's limitations. Namely, by extending the time-filtering procedure of TNT to ensemble architectures such as Random Forests \cite{Breiman2001}, so that TNT's filtering methodology can be compared to state-of-the-art stream learning algorithms such as Adaptive Random and Aggregated Mondrian Forests \cite{GomesBifetRead2017, MourtadaGaiffasScornet2021}, Streaming Random Patches \cite{GomesReadBifet2019} and Streaming Gradient Boosted Trees \cite{GunasekaraPfahringerGomes2024}. We also seek to reduce the sensitivity of TNT on the time filtering threshold $\lambda$ by making it context sensitive. Intuition says that a Hoeffding-like bound may be useful as is the case in incremental decision trees \cite{HultenSpencerDomingos2001}.




\bibliographystyle{abbrv}
\bibliography{references}


\appendix

\section{Appendix} \label{sect:appendix}

The Appendix contains additional details that were not included in the main paper for reasons such as page limits or because they reduced the readability of the main body of text. The Appendix contains the following: The pseudocode for the most critical components of TNT (Section \ref{sect:app_pseudocode}), an empirical evaluation that shows that the Mutual Information of the entire stream decreases as the number of distinct drift events increases (Section \ref{sect:app_MI_EMP}), additional experimental details that ensure paper reproducibility (Section \ref{sect:app_exp_details}), a description of the \verb|RotatingMoons| artificial data stream used to parameter tune TNT (Section \ref{sect:app_rot_moons}), additional details and results from the parameter tuning process (Section \ref{sect:app_param_tune}), additional results (Section \ref{sect:app_extra_results}), and a postmortem that reflects on the main contributions of this paper (Section \ref{sect:app_postmortem}).



\subsection{TNT Pseudocode} \label{sect:app_pseudocode}

This section includes pseudocode for the most critical components of TNT. They are, the impurity reduction (learning) procedure (Alg. \ref{alg:tnt_build}) and the time filtering (Alg. \ref{alg:tnt_temporal_filter}) processes.

\begin{algorithm}[ht]
\caption{Time Neutralising Decision Tree Construction}\label{alg:tnt_build}
\begin{algorithmic}[1]
\Require Features $X$, Labels $y$, Time indices $t$
\Require Hyperparameters: Min samples $N_{min}$, Min info gain $\tau$, Time penalty $\lambda$, MWU alpha $\alpha_{mwu}$

\Function{BuildTree}{$X, y, t$}
    \State $N \gets |y|$
    \If{$N < N_{min}$ \textbf{or} $\text{IsPure}(y)$}
        \State \Return $\text{LeafNode}(\text{mode}(y))$
    \EndIf

    \State $\Delta_{s}, f^*, v^* \gets \Call{BestSpatialSplit}{X, y}$
    \State $\Delta_{t}, t_{split} \gets \Call{BestTemporalSplit}{t, y}$

    \If{$\Delta_{t} > (\Delta_{s} + \lambda)$}
        \State $X_{f}, y_{f}, t_{f} \gets \Call{TemporalFilter}{X, y, t, \alpha_{mwu}, N_{min}, \tau}$ \Comment{See Alg. \ref{alg:tnt_temporal_filter}}
        
        \If{$|y_f| < N_{min}$ \textbf{or} $\text{IsPure}(y_f)$}
            \State \Return $\text{LeafNode}(\text{mode}(y_f))$
        \EndIf
        
        \State $\Delta_{s}, f^*, v^* \gets \Call{BestSpatialSplit}{X_{f}, y_{f}}$ \Comment{Re-evaluate spatial split}
        \State $X, y, t \gets X_{f}, y_{f}, t_{f}$ \Comment{Update reference data for downstream nodes}
    \EndIf

    \State $\text{Node} \gets \text{SplitNode}(f^*, v^*)$ \Comment{Create new split decision node}
    \State $\text{Mask}_{L} \gets X[:, f^*] \leq v^*$
    
    \State $\text{Node.Left} \gets \Call{BuildTree}{X[\text{Mask}_{L}], y[\text{Mask}_{L}], t[\text{Mask}_{L}]}$
    \State $\text{Node.Right} \gets \Call{BuildTree}{X[\neg\text{Mask}_{L}], y[\neg\text{Mask}_{L}], t[\neg\text{Mask}_{L}]}$

    \State \Return Node
\EndFunction
\end{algorithmic}
\end{algorithm}

\begin{algorithm}[ht]
\caption{TNT Temporal Filtering. For the leaf node constructors, we use python-like notation where $t$, $y$ $\min$ and $\max$ refer keyword arguments that set properties of the same name.}\label{alg:tnt_temporal_filter}
\begin{algorithmic}[1]
\Require Features $X$, Labels $y$, Time indices $t$
\Require Hyperparameters: MWU alpha $\alpha_{mwu}$, Min samples $N_{min}$, Min Info Gain $\tau$

\Function{GetTimeLeaves}{$t, y$}
    \If{$|y| < N_{min}$ \textbf{or} $\text{IsPure}(y)$}
        \State \Return $\text{LeafNode}(t=t, y=y, \min=\min(t), \max=\max(t))$
    \EndIf
    
    \State $\Delta, t_{split} \gets \Call{BestTemporalSplit}{t, y}$
    \If{$\Delta \leq \tau$} \Comment{Create leaf if split is not informative enough.}
        \State \Return $\text{LeafNode}(t=t, y=y, \min=\min(t), \max=\max(t))$
    \EndIf
    
    \State $\text{Mask}_{L} \gets t \leq t_{split}$
    \State $Leaves_L \gets \Call{GetTimeLeaves}{t[\text{Mask}_{L}], y[\text{Mask}_{L}]}$
    \State $Leaves_R \gets \Call{GetTimeLeaves}{t[\neg\text{Mask}_{L}], y[\neg\text{Mask}_{L}]}$
    
    \State \Return $Leaves_L \cup Leaves_R$
\EndFunction

\State
\Function{TemporalFilter}{$X, y, t, \alpha_{mwu}, N_{min}, \tau$}
    \State $Leaves \gets \Call{GetTimeLeaves}{t, y}$
    \State $\Call{sort}{Leaves}$ \Comment{Sort leaves in ascending time order}
    
    \State $L_{recent} \gets Leaves[-1]$ \Comment{Get leaf that represents most recent window of data}
    \State $y_{recent} \gets L_{recent}.y$
    \State $Mask_{keep} \gets (t \geq L_{recent}.\min)$

    \For{$L_{hist} \textbf{ in } Leaves[0, 1, \dots, -2]$} \Comment{-2 refers to the second last item in the list}
        \State $p \gets \Call{MannWhitneyU}{y_{recent}, L_{hist}.y}$
        
        \If{$p > \alpha_{mwu}$} \Comment{Distributions are similar so we can merge the leaves.}
            \State $Mask_{keep} \gets Mask_{keep} \lor (t \geq L_{hist}.\min \textbf{ and } t \leq L_{hist}.\max)$
        \EndIf
    \EndFor

    \State \Return $X[Mask_{keep}], y[Mask_{keep}], t[Mask_{keep}]$
\EndFunction
\end{algorithmic}
\end{algorithm}

\subsection{Mutual Information decreases as number of distinct drift events increases.} \label{sect:app_MI_EMP}

In Section \ref{sect:method_motive} we noted that the number of \emph{distinct} concept drift events in some feature subspace $|\mathcal{D}| \to \infty$, the global Mutual Information of that subspace tend to 0.0:

\begin{equation}
    \lim_{|\mathcal{D}| \to \infty} I_{\bar{P}}(X;Y) = 0.0
\end{equation}

\begin{figure}[ht]
  \centering
  \includegraphics[width=\linewidth]{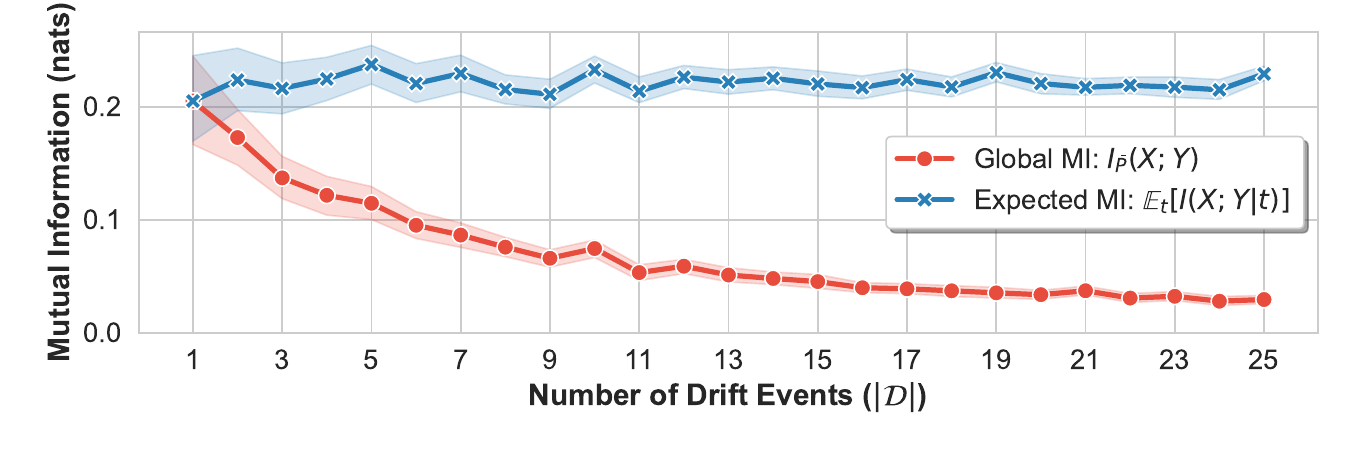}
  \caption{shows how global Mutual Information (MI) (red) in a univariate stream collapses as the number of distinct drift events $|D|$ ($K$ in the main paper) $\to \infty$ while the expected MI of all distinct concepts (blue) remains high. This Fig. shows why active forgetting is useful. By only looking at the most recent concept in a subspace of the stream, that subspace remains informative (i.e. predictive).}
  \label{fig:app_mi_collapse}
\end{figure}

Intuitively, this should make sense. If a \emph{distinct} concept is defined as a concept that is not the same as any previously encountered concept, then as $|\mathcal{D}| \to \infty$, the feature space is reduced to random noise when viewing the entire subspace throughout the entire lifetime of the stream. Random noise by definition contains no Mutual Information. This sections provides empirical evidence in support of this claim. We do this by constructing a 100 synthetic univariate streams for a binary classification task where each class is sampled from a Gaussian distribution with means $\mu_1$ $\mu_2$ resp. with unit variance. Every 50 instances we randomly alter the means of these Gaussians simulating an abrupt drift. We do this for 25 drift events recording the global Mutual Information of the entire stream as well as the average Mutual Information of the each concept after every drift event. Fig. \ref{fig:app_mi_collapse} shows the results of this process whereby we can see that while the average Mutual Information of each Concept remains high (blue line), the global Mutual Information of the entire stream tends to 0.0. It is this phenomena that inspired the creation of TNT. By using time as a feature, we can extract the individual concepts and filter out old ones, increasing the Mutual Information of previously uninformative subspaces. The code to generate Fig. \ref{fig:app_mi_collapse} has been included with the TNT source code.

\subsection{Experimental Details} \label{sect:app_exp_details}

This Section provides additional experimental details. Firstly, Table \ref{tab:datastreams} includes a summary of the number features, class, and instances for every data stream used in this work. For detailed descriptions on each data stream, we refer the reader to the accompanying citations. As noted in Section \ref{sect:exp_design}, we acquired these data streams from the USP DS repository\footnote{\url{https://sites.google.com/view/uspdsrepository}} \cite{SouzaReisMaletzke2020}. All streams are stored as CSV files. During preprocessing, we found that the \verb|Asfault|, \verb|Posture| and \verb|SmartMeter| streams were malformed. Further investigation revealed that this was due to a formatting error whereby two samples were stored as a single entry. These malformed entries were simply deleted. Furthermore, the feature 'at3' in the \verb|Luxembourg| stream was revealed to be overly informative (i.e., it perfectly predicts the class label). We did not include it when running experiments. Before running experiments all numerical features in all streams are standardised. Although this step is unnecessary as we only investigated Decision Tree architectures. We do also want to note that the listed features, and instance counts of the downloaded CSVs do not always match the values listed on the USP DS repository website$^1$. We do not know why this is the case but Table \ref{tab:datastreams} includes our updated counts. Furthermore, undocumented behaviour in the Kappa statistic calculated by river results in a value 0.0 being returned if only one class exists in both the labels and predictions. This can occur when there is only one class present over some window in a autocorrelated or imbalanced stream. We fixed this post-hoc but setting the Kappa statistic to 1.0 for all windows where an algorithm had a 0.0 Kappa score, but perfect Balanced Accuracy and Accuracy score. This does not meaningfully change the results in any way, but is a more accurate reflection of the performance of the algorithms evaluated.

\begin{figure}[ht]
  \centering
  \includegraphics[width=\linewidth]{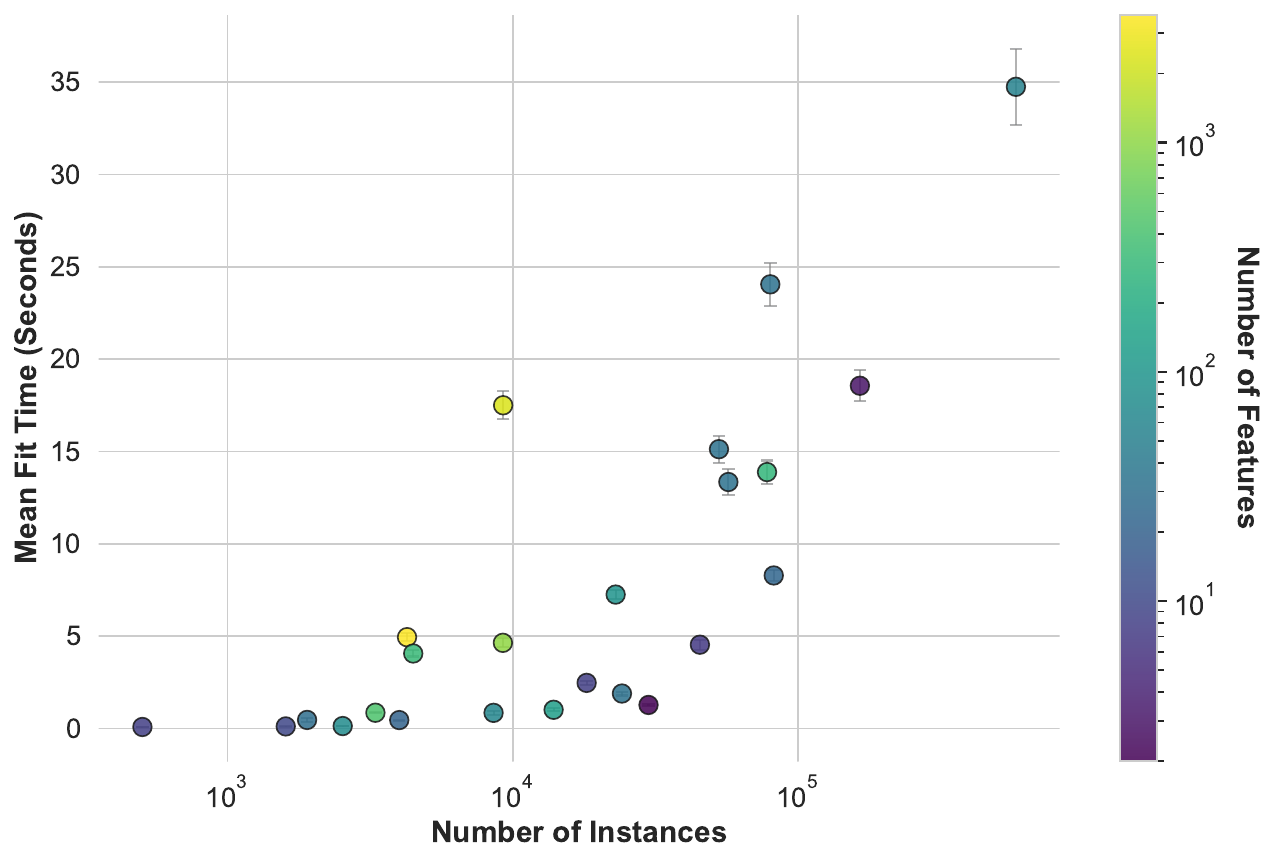}
  \caption{shows the compute time (in seconds) taken for TNT to fit each stream (Table \ref{tab:datastreams}) evaluated. Each stream (point) is plotted by number of instances (x-axis) and number of features (colour bar). Error bars indicate the standard deviation over 10 replicates. Note that this plot shows the time it takes to fit the entire stream, and is not indicative of the time taken to evaluate TNT as described in Section \ref{sect:exp_design} where TNT is retrained every 50 instances on an ever growing stream. This plot only serves to show the training compute trends as both the number of instances and features increase.}
  \label{fig:app_TNT_time_scale}
\end{figure}

All code and experiments were run using \verb|Python 3| with the Batch Decision Trees sourced from \verb|scikit-learn| \cite{PedgregosaVaroquauxGramfort2011} and Stream Learning Decision Trees from \verb|river| \cite{Montiel2021}. The exception to this is PLASTIC \cite{HeydenGomesFouche2024} which does not have an implementation in \verb|river| so instead we used the \verb|CapyMOA|. Note that we chose not to use \verb|CapyMOA| for the other Stream Learning Decision Trees because the package is still very much under active development at the time of writing. TNT is implemented in \verb|Python 3|, with time consuming operations written in \verb|Cython|. TNT is \verb|scikit-learn| compatible. All experiments were run using Google Colab or a 2020 MacBook Pro with 2 GHz Quad-Core Intel i5 and 16GB RAM.
The average run-time for an experiment is highly algorithm and stream-dependent. For example, TNT run on the \verb|Forest Cover Type| stream took $\sim$50 hours, but it takes less than a minute to run TNT on the \verb|Keystroke| stream. We have included the run time complexity of TNT when trained on the entirety of each stream over 10 replicates in Figure \ref{fig:app_TNT_time_scale}.

For parameter selection, we opted to use default parameters for most models as they are informed by decades of research. For the Batch Decision Trees, we set \verb|min_samples_leaf| = 10 as the parameter to balance under/overfitting. The logic being that it is half the $N_{min}$ used by TNT as so they approximately equivalent when taking implementation specifics into account. For the Stream Learners, we used the default parameters for all \verb|river| models. For PLASTIC, we set \verb|split_confidence| to $1^{-3}$ and unbounded the \verb|max_depth| parameter as in the original implementation\footnote{\url{https://github.com/heymarco/CapyMOA-PLASTIC/blob/plastic/src/capymoa/learner/classifier/plastic.py}}. We also set the \verb|leaf_prediction| strategy to Majority Class (MC) as the other options did not appear to be working correctly and would frequently return \verb|None| even after PLASTIC trained on several hundred instances.

All experiments run in this work are stored as CSV files, and all figures are generated from these files. The source code contains a summary description of the structure of these CSVs. For statistical testing and Quasi-Monte Carlo Latin Hypercube sampling, we make use of \verb|scipy| \cite{VirtanenGommersOliphant2020}, and use $p=0.05$ for all statistical tests performed. For constructing the Critical Difference diagram (Fig. \ref{fig:cd_diagram} in the main paper), we use \verb|scikit-posthocs|\footnote{\url{https://scikit-posthocs.readthedocs.io/en/latest/index.html}}. As noted before, we made use of pseudo-random number generators to ensure reproducibility. In general, a default seed value of 42 was used. For the 10 replicates in the main evaluation, we used the following seeds that were initially themselves randomly generated: [1341, 1967, 2414, 3258, 3485, 3676, 5588, 5616, 7079, 8172].

For some of the larger streams, compute limitations dictated that we downsample the training set for DT$_\infty$ and TNT. They are: \verb|Forest Cover Type| and \verb|Posture| for both models and \verb|Heartbeats| for just DT$_\infty$. When the total stream size became larger than 10 000, we performed stratified sampling (of 10 000 instances) to preserve class label distributions. This process was also psuedo-random ensuring that it was equivalent across both methods.

\begin{table}[ht]
    \centering
    \caption{The USP DS streams \cite{SouzaReisMaletzke2020} evaluated in the work.}
    \label{tab:datastreams}
    \begin{tabular}{l|c|c|c}
         \toprule
         Name & Features & Classes & Instances \\
         \midrule
         \rowcolor[gray]{0.9}
         Asfault (AF) \cite{Souza2018} & 64 & 5 & 8,565  \\
         Cancer (CA) \cite{SouzaParmezanChowdhury2021} & 2,352 & 7 & 9,256  \\
         \rowcolor[gray]{0.9}
         Chess (CH) \cite{Zliobaite2011} & 8 & 3 & 503  \\
         Electricity (EL) \cite{Harries1999} & 8 & 2 & 45,312  \\
         \rowcolor[gray]{0.9}
         Forest Cover Type (FC) \cite{BlackardDean1999} & 54 & 7 & 581,012  \\
         Gas Sensor Array (GS) \cite{VergaraVembuAyhan2012} & 128 & 6 & 13,910  \\
         \rowcolor[gray]{0.9}
         Heartbeats (HB) \cite{UlanovaBegumShokoohi2016} & 280 & 8 & 77,904  \\
         Insects-Abrupt (IA) \cite{SouzaReisMaletzke2020} & 33 & 6 & 52,848  \\
         \rowcolor[gray]{0.9}
         Insects-Gradual (IG) \cite{SouzaReisMaletzke2020} & 33 & 6 & 24,150  \\
         Insects-Incremental (II) \cite{SouzaReisMaletzke2020} & 33 & 6 & 57,018  \\
         \rowcolor[gray]{0.9}
         Insects-Reoccurring (IR) \cite{SouzaReisMaletzke2020} & 33 & 6 & 79,986  \\
         Keystroke (KS) \cite{SouzaSilvaGamaBatista2015} & 10 & 4 & 1,600  \\
         \rowcolor[gray]{0.9}
         Luxembourg (LX) \cite{Zliobaite2011} & 30 & 2 & 1,901  \\
         MIRS (MS) \cite{DauKeoghKamgar2018} & 3,600 & 2 & 4,260 \\
         \rowcolor[gray]{0.9}
         NOAA Weather (NW) \cite{DitzlerPolikar2012} & 8 & 2 & 18,159  \\
         Outdoor Objects (OD) \cite{LosingHammerWersing2015} & 21 & 40 & 4000  \\
         \rowcolor[gray]{0.9}
         Ozone (OZ) \cite{DheeruTaniskidou2017} & 72 & 2 & 2,534  \\
         Posture (PO) \cite{ReisFlachMatwin2016fast} & 3 & 11 & 164,859  \\
         \rowcolor[gray]{0.9}
         Powersupply (PS) \cite{Zhu2010}  & 2 & 24 & 29,928  \\
         Rialto (RT) \cite{LosingHammerWersing2016} & 23 & 10 & 82,250  \\
         \rowcolor[gray]{0.9}
         SmartMeter (SM) \cite{SouzaParmezanChowdhury2021} & 96 & 10 & 22,949  \\
         StarLightCurves (SL) \cite{SouzaParmezanChowdhury2021} & 1024 & 3 & 9,236\\
         \rowcolor[gray]{0.9}
         UWave (UV) \cite{SouzaParmezanChowdhury2021} & 315 & 8 & 4,478  \\
         Yoga (YG) \cite{DauKeoghKamgar2018} & 426 & 2 & 3,300 \\
         \bottomrule
    \end{tabular}
\end{table}

\subsection{Rotating Moons Synthetic Data Stream} \label{sect:app_rot_moons}

\begin{figure}[ht]
  \centering
  \includegraphics[width=\linewidth]{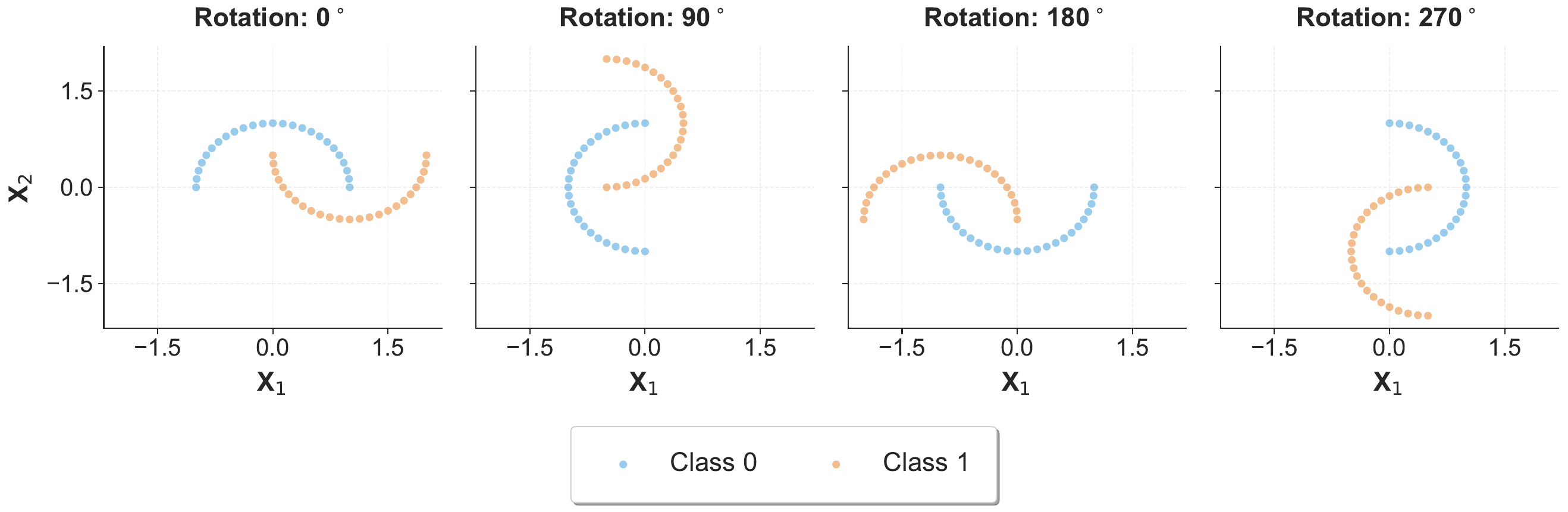}
  \caption{shows a visualisation of the synthetic RotatingMoons data stream using a rotation angle of 90$^\circ$ for one cycle. To simulating reoccurring concepts, the cycle is repeated left $\to$ right.}
  \label{fig:app_rotating_moons}
\end{figure}

In this work we also introduce a new synthetic data stream called \verb|RotatingMoons|. It uses the \verb|make_moons| generator provided by \verb|scikit-learn|\footnote{https://scikit-learn.org/stable/modules/generated/sklearn.datasets.make\_moons.html} which is a two-dimensional non-linear binary classification task. \verb|RotatingMoons| works by accepting a rotation $R$, and samples per cycle $n$ parameters. By generating $n$ samples from \verb|make_moons|, a rotation is applied to the data: 

\begin{equation}
    X_{\theta} = 
    \begin{bmatrix}
        x_1^1 & x_2^1 \\
        x_1^2 & x_2^2 \\
        \ldots & \ldots \\
        x_1^n & x_2^n
    \end{bmatrix} 
    \begin{bmatrix}
        \cos(\theta) & \sin(\theta) \\
        -\sin(\theta) & \cos(\theta) \\
    \end{bmatrix}
\end{equation}

where $\theta$ is the rotation angle in radians determined by ever increasing multiples of $R$ until a full revolution is complete. To simulate reoccurring concepts, this process can be repeated infinitely. Fig. \ref{fig:app_rotating_moons} visualises the \verb|RotatingMoons| stream at different rotations and the code for the generator is included with the Supplementary Materials.

\subsection{Parameter Tuning} \label{sect:app_param_tune}

Recall from Section \ref{sect:exp_design_param_tune} that TNT has 5 tunable parameters: filter factor $\lambda$, minimum split node samples $N_{min}$, minimum time node entropy split $\tau$, confidence threshold for determining time node merges $p_{mwu}$, and the maximum number of threshold splits to consider per feature $\text{thresholds}_{max}$. To avoid data leakage and bias \cite{SouzaReisMaletzke2020}, we parameter tune TNT on 5 synthetic streams which cover a variety of drift types. They are: \verb|Hyperplane| \cite{HultenSpencerDomingos2001}, \verb|RandomRBFDrift| \cite{Montiel2021}, \verb|SEA| \cite{StreetKim2001}, \verb|STAGGER| \cite{SchlimmerGranger1986} and a custom stream \verb|RotatingMoons| (\ref{sect:app_rot_moons}). After some initial exploration of the hyper-parameters, we used Quasi-Monte Carlo Latin Hypercube Sampling \cite{Niederreiter1992} to generate $100$ parameter configurations which we explored across all synthetic streams, recording the final accuracy of the model per configuration per stream. For each stream, we then train a Random Forest \cite{Breiman2001} regressor using the parameter configurations as input and the final accuracy as a target. From each trained Forest, we extract feature importances (Section \ref{sect:app_param_tune_fi}). We also plot each parameter and obtained accuracy per stream using LOWESS regression to analyse per feature level dependencies (Section \ref{sect:app_param_tune_sa}). Lastly, we also ran a leaf node ablation (Section \ref{sect:app_param_tune_la}). Table \ref{tab:app_hyperparameters} notes the range explored for each parameter configuration, and Table \ref{tab:app_synth_generator_params} summarises the parameter values used to generate each artificial stream.

\begin{table}
  \caption{Hyper-parameters, the range of values explored, and their respective final values used in our evaluation.}
  \label{tab:app_hyperparameters}
  \centering
  \begin{tabular}{l|l|l}
    \toprule
    Name     & Range Explored    & Final Value \\
    \midrule
    \rowcolor[gray]{0.9}
    $\lambda$ & [0.01, 0.5] & 0.20 \\
    $N_{min}$ & [10, 50] & 20 \\
    \rowcolor[gray]{0.9}
    $\tau$ & [0.01, 0.1] & 0.05 \\
    $\alpha_{mwu}$ & [0.0001, 0.1] & 0.05 \\
    \rowcolor[gray]{0.9}
    thresholds$_{max}$ & [5, 25] & 15 \\
    leaf model & \{MC, LR, NB\} & LR \\
    \bottomrule
  \end{tabular}
\end{table}

\begin{table}
  \caption{Synthetic generators used in this work, their parameters, and total number of instances (N). Random seed 42 is used for all streams.}
  \label{tab:app_synth_generator_params}
  \centering
  \begin{tabular}{l|l|l}
    \toprule
    Name     & Range Explored    & Total (N) \\
    \midrule
    \rowcolor[gray]{0.9}
    Hyperplane \cite{HultenSpencerDomingos2001} & mag\_change = 0.2 & 5000 \\
    RandomRBFDrift \cite{Montiel2021} & change\_speed = 1.0 & 1000 \\
    \rowcolor[gray]{0.9}
    & sample\_per\_phase = 100 &  \\
    \rowcolor[gray]{0.9}
    RotatingMoons (\ref{sect:app_rot_moons}) & rotation\_angle = 90 & 1600 \\
    \rowcolor[gray]{0.9}
    & cycles = 4 &  \\
    SEA \cite{StreetKim2001} & sample\_per\_variant = 200 & 800 \\
    \rowcolor[gray]{0.9}
    STAGGER \cite{SchlimmerGranger1986} & sample\_per\_variant = 150 & 450 \\
    \bottomrule
  \end{tabular}
\end{table}

\subsubsection{Feature Importance} \label{sect:app_param_tune_fi}

\begin{figure}[ht]
     \centering
     \begin{subfigure}{0.7\textwidth}
         \centering
         \includegraphics[width=\textwidth]{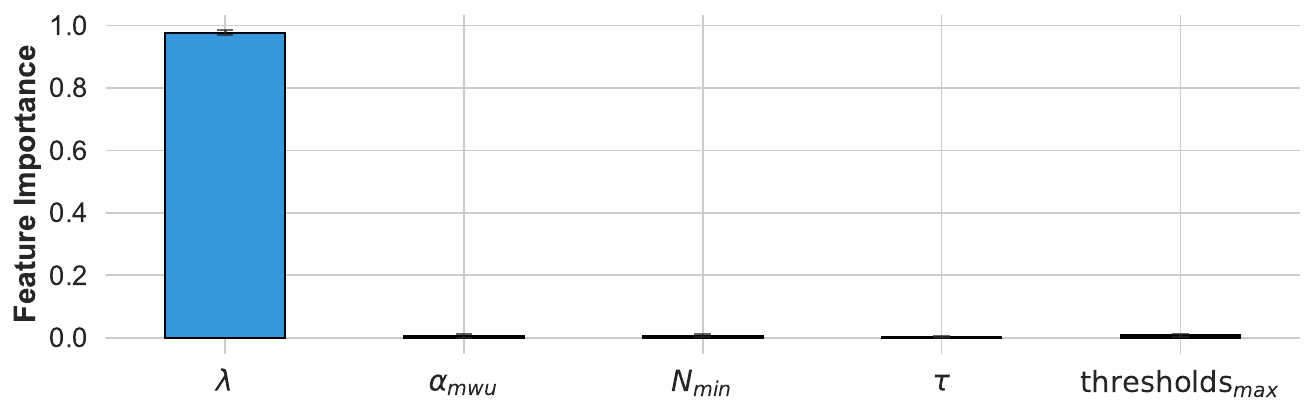}
        \caption{Hyperplane}
        \label{fig:app_hyperplane_fi}
     \end{subfigure}
     \begin{subfigure}{0.7\textwidth}
         \centering
         \includegraphics[width=\textwidth]{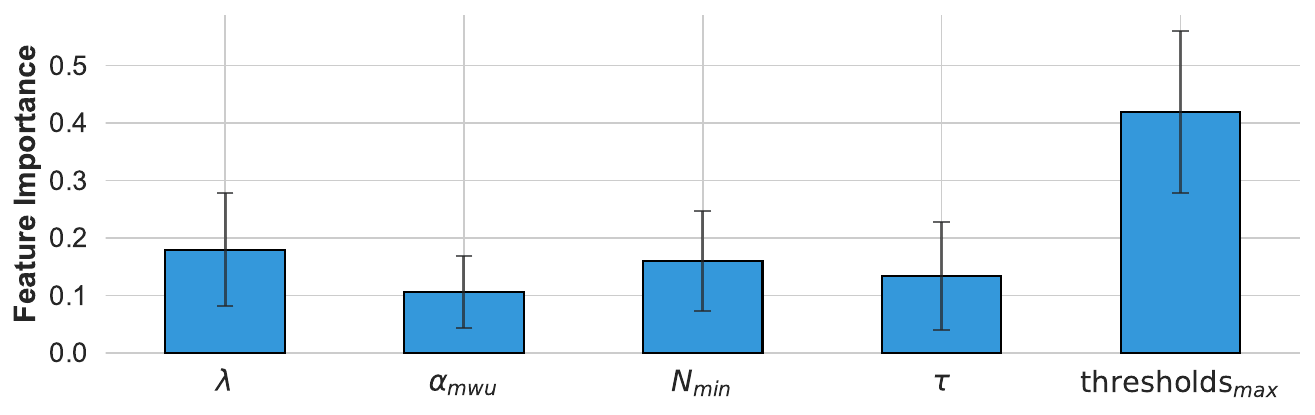}
        \caption{RandomRBFDrift}
        \label{fig:app_randomrbf_fi}
     \end{subfigure}
     \begin{subfigure}{0.7\textwidth}
         \centering
         \includegraphics[width=\textwidth]{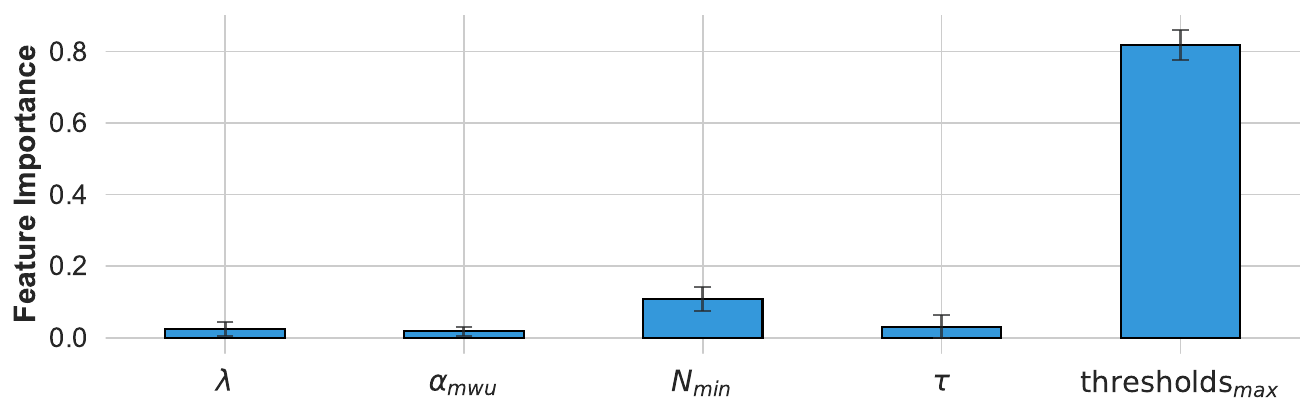}
        \caption{RotatingMoons}
        \label{fig:app_rotating_moons_fi}
     \end{subfigure}
     \begin{subfigure}{0.7\textwidth}
         \centering
         \includegraphics[width=\textwidth]{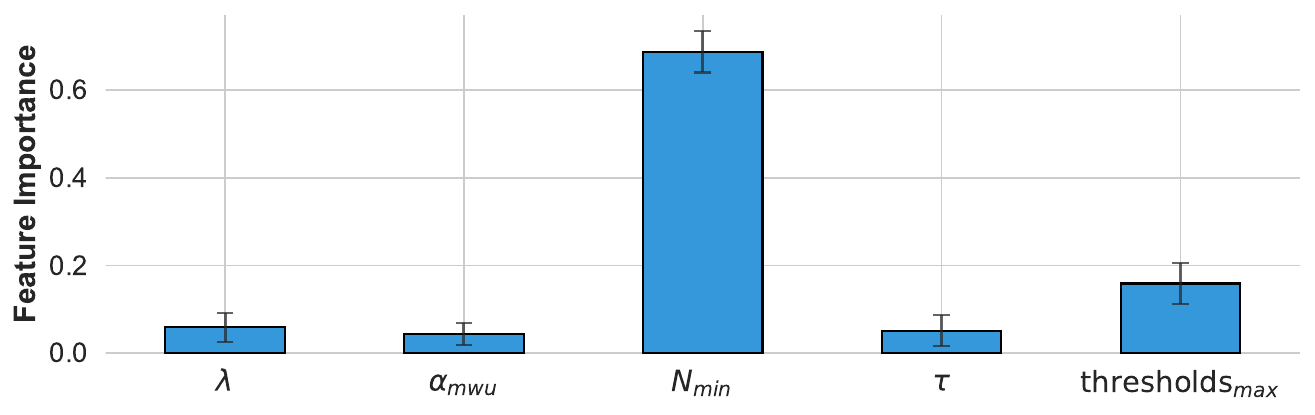}
        \caption{SEA}
        \label{fig:app_sea_fi}
     \end{subfigure}
     \begin{subfigure}{0.7\textwidth}
         \centering
         \includegraphics[width=\textwidth]{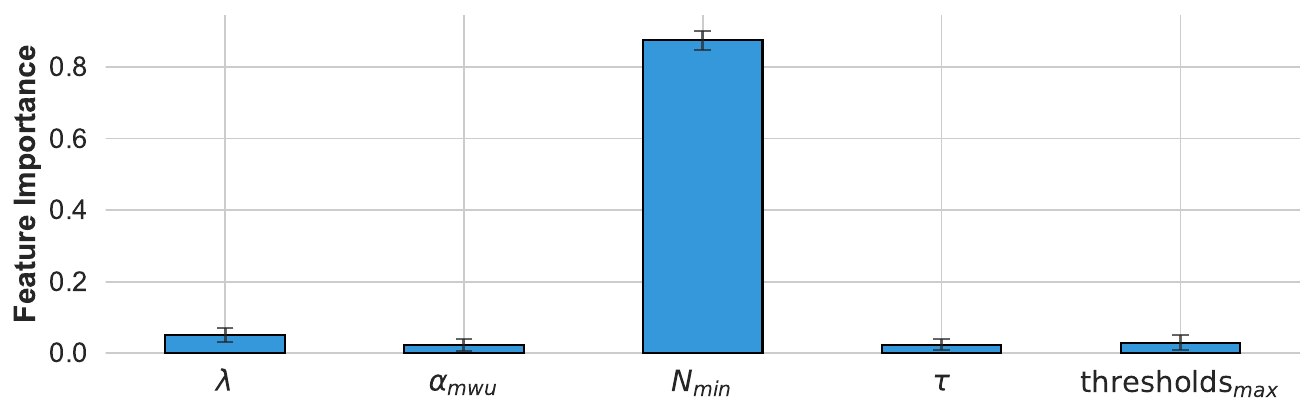}
        \caption{STAGGER}
        \label{fig:app_stagger_fi}
     \end{subfigure}
     \caption{Feature Importances (Mean Decrease in Impurity) of a Random Forest Regressor trained to predict the accuracy of various parameter configurations of TNT per synthetic stream investigated.}
     \label{fig:app_fi}
\end{figure}

Figure \ref{fig:app_fi} shows the result of the feature importance analysis in the form of mean impurity reduction. A Random Forest regressor is trained using parameter configurations generated from the Latin Hypercube Sampling using the parameter ranges in Table \ref{tab:app_hyperparameters} as input features $X$ with the target $y$ being the accuracy that TNT achieved with that parameter configuration. This is done per synthetic stream. Results show that across most synthetic streams investigated, the minimum node split size $N_{min}$ and number of spatial splits considered $\text{thresholds}_{max}$ are the most important features. This makes sense as they control how effectively TNT is able to generalise in a given domain. Just like in regular batch decision trees, too much splitting causes a tree to overfit while not enough splitting causes a tree to underfit. Interestingly, \verb|Hyperplane| is the only synthetic stream where the forgetting factor $\lambda$ is a significant predictor of accuracy. In \verb|Hyperplane|, the decision boundary $P(Y|X)$ is constantly moving while the covariates $P(X)$ remain stable. This is precisely the setting where active forgetting is useful (Section \ref{sect:method_motive}), and we see that setting a forgetting factor $\lambda$ as low as possible allows TNT to effectively filter out old irrelevant instances thus adapting to concept drift.

\subsubsection{Per Feature Sensitivity Analysis} \label{sect:app_param_tune_sa}

\begin{figure}[ht]
     \centering
     \begin{subfigure}{\textwidth}
         \centering
         \includegraphics[width=\textwidth]{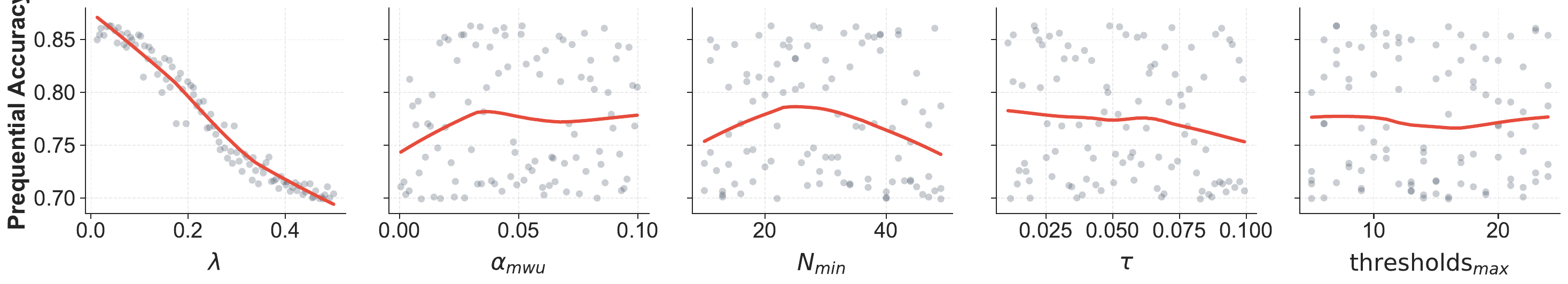}
        \caption{Hyperplane}
        \label{fig:app_hyperplane_sa}
     \end{subfigure}
     \begin{subfigure}{\textwidth}
         \centering
         \includegraphics[width=\textwidth]{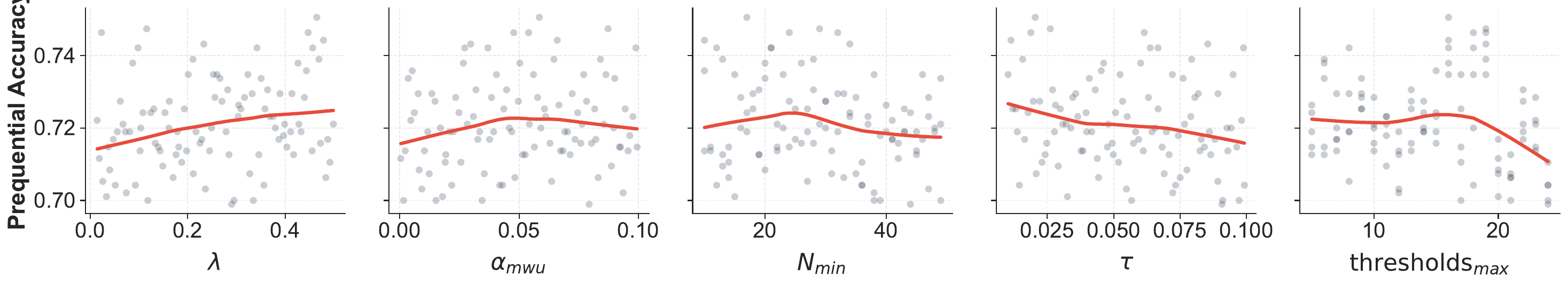}
        \caption{RandomRBFDrift}
        \label{fig:app_randomrbf_sa}
     \end{subfigure}
     \begin{subfigure}{\textwidth}
         \centering
         \includegraphics[width=\textwidth]{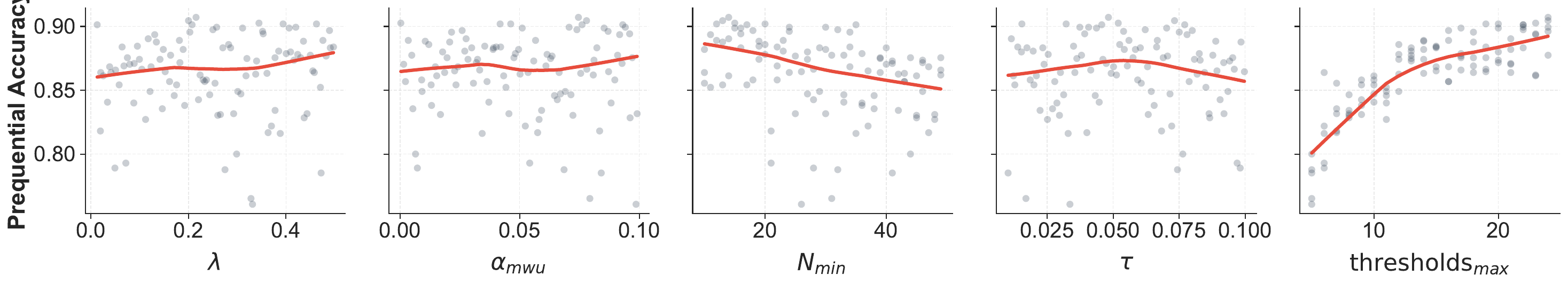}
        \caption{RotatingMoons}
        \label{fig:app_rotating_moons_sa}
     \end{subfigure}
     \begin{subfigure}{\textwidth}
         \centering
         \includegraphics[width=\textwidth]{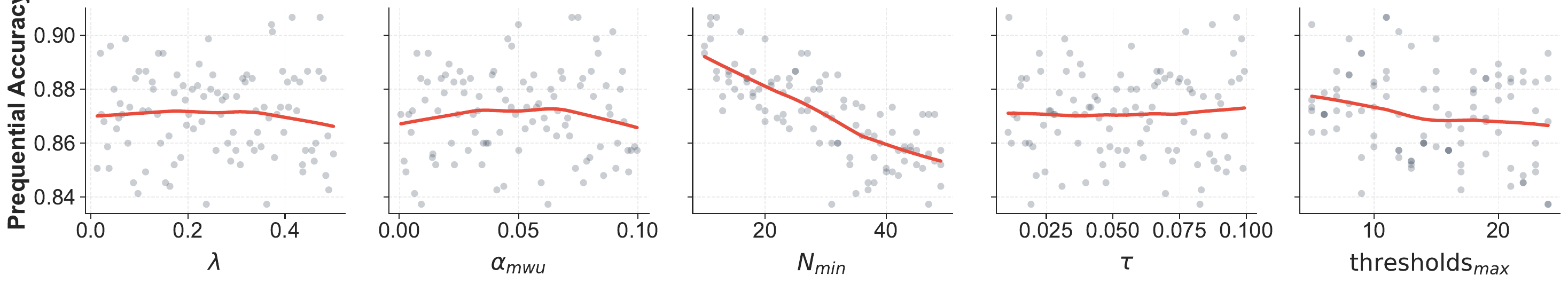}
        \caption{SEA}
        \label{fig:app_sea_sa}
     \end{subfigure}
     \begin{subfigure}{\textwidth}
         \centering
         \includegraphics[width=\textwidth]{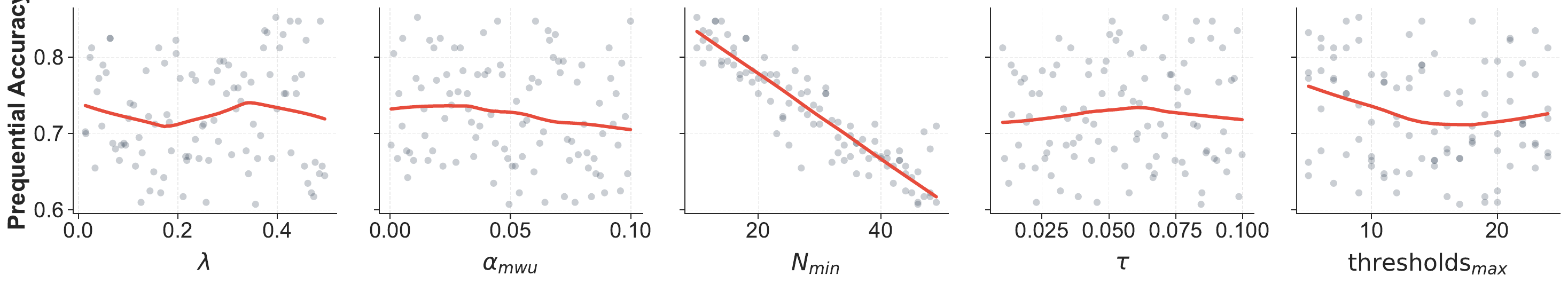}
        \caption{STAGGER}
        \label{fig:app_stagger_sa}
     \end{subfigure}
     \caption{Per Feature Sensitivity Analysis of each of TNT's hyper-parameters per synthetic stream investigated. Trend lines (red) are determined by LOWESS regression.}
     \label{fig:app_sa}
\end{figure}

While feature importance extraction allows one to understand the complex interactions between hyper-parameters, we also performed sensitivity analysis on a per-feature level per stream using LOWESS regression. Figure \ref{fig:app_sa} shows the results which reinforces the findings of the feature importance analysis. $N_{min}$ is strongly correlated with the accuracy of TNT on the \verb|SEA| and \verb|STAGGER| streams while $\text{thresholds}_{max}$ strongly correlates with accuracy on the \verb|RotatingMoons| stream. Interestingly, this per-feature perspective on the \verb|RandomRBFDrift| stream implies that no single feature overwhelmingly predicts TNT's accuracy. This is echoed in Fig. \ref{fig:app_randomrbf_fi} where all features carry some predictive power. Filter factor $\lambda$ negatively correlates with accuracy for \verb|Hyperplane| stream for reasons described in the previous Section.

\subsubsection{Leaf Node Ablation} \label{sect:app_param_tune_la}

Incremental Decision Trees, such as Hoeffding Trees \cite{HultenSpencerDomingos2001}, often place Naive Bayes classifiers on their leaf nodes. They do this to mitigate underfitting while the Hoeffding Tree conservatively learns split decisions over the lifetime of a stream. We investigated if this type of design (replacing leaf impure leaf nodes with lightweight classifiers) would benefit TNT and make it more robust to underfitting. To do this, we performed an ablation study using the same evaluation procedure detailed in Section \ref{sect:exp_design}. To avoid data leakage, we evaluated these ablations on the synthetic streams (Table \ref{tab:app_synth_generator_params}) and not on the USP DS repository. We investigated three versions of TNT: (1) a baseline where TNT uses Majority class prediction on its leaf nodes, (2) TNT where impure leaf nodes are replaced by a Logistic Regression (LR) model and (3) TNT where impure leaf nodes are replaced by a Naive Bayes (NB) model.
For LR we set a maximum of 200 iterations to reach convergence and a random seed of 42 for every new instantiation to ensure reproducibility. In practice, it may be worth increasing the maximum iterations and exploring different random states. Results of this evaluation are summarised in Table \ref{tab:app_leaf_ablation} where there is clear evidence that using a lightweight classifier on impure leaf nodes is beneficial. Overall, there wasn't a significant difference when comparing LR to NB. However, given that average Kappa score of the LR variant was greatest (by a small margin) amongst all synthetic streams investigated, we chose to use it for our main evaluation. 

\begin{table}[ht]
    \centering
    \caption{Mean Kappa scores for different variants of TNT. MC uses Majority Class to make predictions at leaf nodes, while LR and NB replace leaf nodes with Logistic Regression and Naive Bayes models resp. Bold values indicate highest average Kappa score.}
    \label{tab:app_leaf_ablation}
    \begin{tabular}{l|c|c|c}
    \toprule
        \multicolumn{1}{c|}{} & \textbf{MC} & \textbf{LR} & \textbf{NB}  \\
        \midrule
        \rowcolor[gray]{0.9}
        \textbf{Hyperplane} & \textbf{0.61} & \textbf{0.61} & \textbf{0.61} \\
        \textbf{RandomRBFDrift} & 0.40 & \textbf{0.44} & \textbf{0.44} \\
        \rowcolor[gray]{0.9}
        \textbf{RotatingMoons} & 0.78 & 0.81 & \textbf{0.82} \\
        \textbf{SEA} & 0.78 & \textbf{0.81} & 0.80 \\
        \rowcolor[gray]{0.9}
        \textbf{STAGGER} & 0.44 & \textbf{0.57} & 0.53 \\
        \midrule
        \textbf{Average} & 0.60 & \textbf{0.65} & 0.64 \\ 
        \bottomrule
    \end{tabular}
\end{table}

\subsection{Additional Results} \label{sect:app_extra_results}

This section includes additional results that did not make it into the main paper. Fig. \ref{fig:cd_diagram} shows the Critical Difference diagram of the methods evaluated across all streams and Table \ref{tab:results_main_ranks} lists the ranking per stream for all methods investigated in this work. We used fractional ranking whereby similarly performing models were given the same rank. For example, if the Kappa scores for 4 arbitrary models on some arbitrary stream were: [85, 80, 80, 60], they would be given the ranks [1, 2.5, 2.5, 4] resp. Table \ref{tab:results_main_ranks} does not present any meaningful results beyond the main table (Table \ref{tab:results_main}), but we have included it for completeness.

\begin{figure}[ht]
  \centering
  \includegraphics[width=\linewidth]{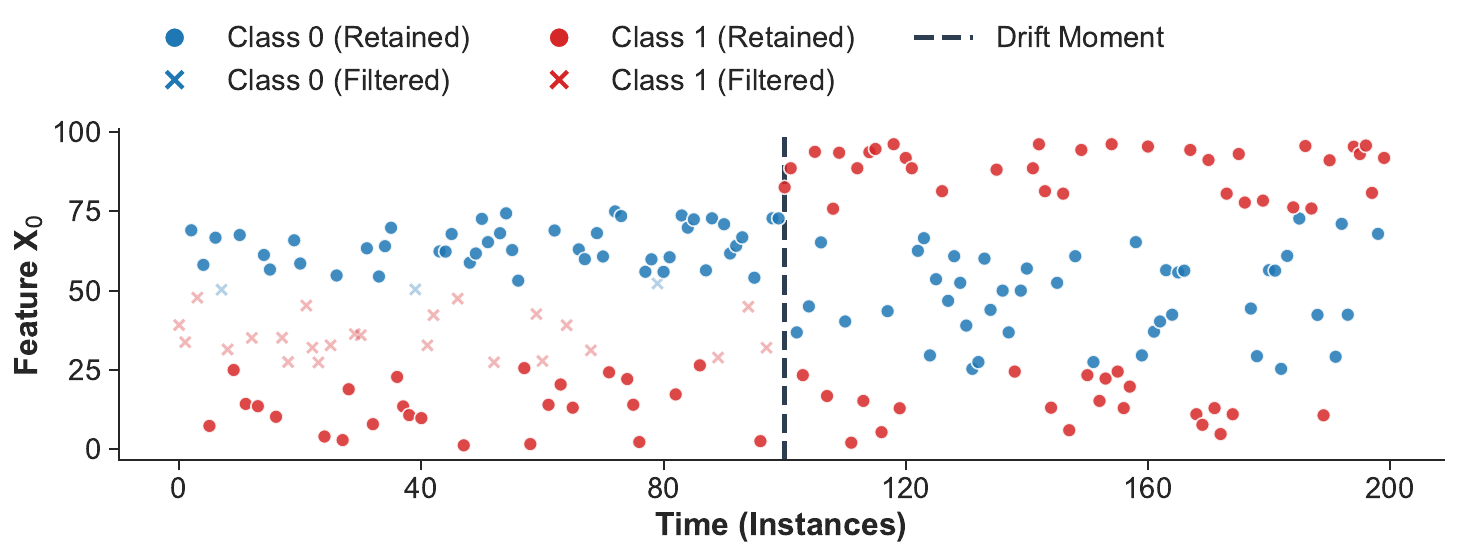}
  \caption{shows how TNT filters instances that are useful to forget. This image was generated by training TNT on a synthetic stream that matches the stream shown in Fig. \ref{fig:TNT_motive} which we used to motivate the design of TNT. TNT filters out instances (crosses) that it deems no longer relevant to the stream after retraining. This Figure matches the original concept in Fig. \ref{fig:TNT_motive} with instances in the X$_{1}=[25, 50]$ subspace (pre-drift event) being filtered out as they are no longer relevant for making predictions once the drift event (dashed line) has occurred.}
  \label{fig:app_TNT_filter}
\end{figure}

We also wanted to show how the forgetting works on an actual data stream, but this very difficult to visualise, and confirm, in high-dimensional spaces. This is because instances that look like they should be excluded when viewing one feature, may be perfectly separable in another feature. Nevertheless, Fig. \ref{fig:app_TNT_filter} showcases the instance filtering of TNT on the same synthetic stream we used to motivate the problem in Section \ref{sect:method_motive}. In Fig. \ref{fig:app_TNT_filter}, TNT is trained on the entire stream and then we track which instances are excluded during the training process. As shown, TNT excluded the instances in the subspace X$_{1}=[25, 50]$ where instances prior to the drift event (dashed line) conflict with new instances post drift event. This confirms that TNT does what we designed it to do: filter out irrelevant instances during training based on the stream's current context, thus enabling TNT to adapt to concept drift.

\begin{landscape}
\begin{table}[htbp]
    \centering
    \caption{Cohen's Kappa for each method (column) and data stream (row). Bolded values indicate the statistically best performing method. Underlined values are used to indicate top performing methods when no single method is best. Mean rank per method is reported with a bold value indicating lowest (best) rank.}
    \label{tab:results_main}
    \begin{adjustbox}{angle=0}
    \begin{tabular}{l|c|c|c|c|c|c|c|c|c|c|c|c|c|c|c|c|c}
    \toprule
        \multicolumn{1}{c|}{} & \multicolumn{2}{c|}{\textbf{Baselines}}  & \multicolumn{5}{c|}{\textbf{Sliding Window}} & \multicolumn{3}{c|}{\textbf{Incremental}}  & \multicolumn{5}{c|}{\textbf{Drift Detectors}} & \multicolumn{2}{c}{\textbf{Ours}} \\
        \cmidrule{2-18}
        \multicolumn{1}{c|}{} & \textbf{LC} & \textbf{MC} & \textbf{DT$_{50}$} & \textbf{DT$_{100}$} & \textbf{DT$_{250}$} & \textbf{DT$_{500}$} & \textbf{DT$_{\infty}$} & \textbf{HT} & \textbf{EFDT} & \textbf{PLS} & \textbf{HAT} & \textbf{LAST} & \textbf{$\text{HT}_{\text{ADWIN}}$} & \textbf{$\text{HT}_{\text{DDM}}$}  & \textbf{$\text{HT}_{\text{D3}}$} & \textbf{TNT} & \textbf{TNT$_{500}$} \\
        \midrule
        \textbf{AF} & 0.19 & 0 & 0.51 & 0.61 & \underline{0.65} & \underline{0.68} & \underline{0.66} & 0.52 & 0.59 & 0.57 & 0.52 & \underline{0.67} & \underline{0.62} & 0.58 & 0.41 & \underline{0.65} & \underline{0.69} \\
        \rowcolor[gray]{0.9}
        \textbf{CA} & -0.01 & 0.01 & 0.07 & 0.11 & 0.14 & 0.17 & \underline{0.20} & 0.03 & 0.03 & 0.02 & 0.03 & 0.09 & 0.04 & 0.03 & 0.02 & \underline{0.18} & 0.17 \\
        \textbf{CH}$^*$ & 0.14 & 0.09 & 0.55 & 0.55 & 0.55 & 0.57 & 0.57 & 0.38 & 0.48 & 0.38 & 0.36 & 0.26 & 0.38 & 0.38 & 0.44 & 0.54 & 0.54 \\
        \rowcolor[gray]{0.9}
        \textbf{EL} & \underline{0.63} & 0 & 0.55 & 0.54 & 0.51 & 0.51 & 0.41 & 0.48 & 0.54 & 0.46 & 0.58 & 0.51 & 0.58 & \underline{0.61} & \underline{0.61} & 0.46 & 0.52 \\
        \textbf{FC} & \textbf{0.84} & 0.13 & 0.18 & 0.18 & 0.28 & 0.53 & 0.27 & 0.42 & 0.48 & 0.48 & 0.45 & 0.55 & 0.40 & 0.57 & 0.48 & 0.27 & 0.56 \\
        \rowcolor[gray]{0.9}
        \textbf{GS} & 0.18 & 0.07 & 0.66 & 0.67 & 0.67 & 0.65 & 0.69 & 0.35 & 0.63 & 0.56 & 0.35 & 0.54 & 0.64 & 0.69 & \textbf{0.76} & 0.75 & 0.69 \\ 
        \textbf{HB} & 0.75 & 0.75 & 0.82 & 0.83 & 0.85 & 0.86 & \textbf{0.88} & 0.53 & 0.70 & 0.86 & 0.74 & 0.84 & 0.81 & 0.82 & 0.80 & 0.85 & 0.82 \\
        \rowcolor[gray]{0.9}
        \textbf{IA} & 0.04 & -0.01 & 0.42 & 0.49 & 0.54 & 0.55 & 0.42 & 0.37 & 0.42 & 0.46 & 0.47 & 0.39 & 0.51 & 0.45 & 0.46 & 0.48 & \textbf{0.58} \\
        \textbf{IG} & 0.11 & -0.01 & 0.47 & 0.53 & 0.57 & 0.59 & 0.42 & 0.37 & 0.39 & 0.41 & 0.39 & 0.37 & 0.42 & 0.42 & 0.48 & 0.55 & \textbf{0.61} \\
        \rowcolor[gray]{0.9}
        \textbf{II}  & -0.02 & -0.06 & 0.25 & 0.36 & 0.42 & 0.45 & 0.41 & 0.41 & 0.42 & \underline{0.47} & 0.44 & 0.43 & \underline{0.47} & 0.45 & 0.46 & 0.45 & 0.46 \\
        \textbf{IR} & 0.09 & 0 & 0.45 & 0.51 & 0.55 & 0.56 & 0.28 & 0.30 & 0.36 & 0.38 & 0.38 & 0.33 & 0.47 & 0.39 & 0.43 & 0.43 & \textbf{0.60} \\
        \rowcolor[gray]{0.9}
        \textbf{KS} & -0.25 & -0.24 & 0.74 & 0.79 & 0.82 & 0.83 & 0.83 & 0.61 & 0.66 & 0.35 & 0.61 & 0.59 & 0.61 & 0.65 & 0.79 & 0.86 & \textbf{0.88} \\
        \textbf{LX}$^*$ & -0.05 & 0 & 0.11 & 0.17 & 0.20 & 0.24 & 0.22 & 0.12 & 0.13 & 0.13 & 0.13 & 0.12 & 0.12 & 0.12 & 0.13 & 0.22 & 0.23 \\
        \rowcolor[gray]{0.9}
        \textbf{MS} & -0.03 & 0 & 0.42 & \underline{0.46} & \underline{0.49} & \underline{0.50} & \underline{0.53} & 0.23 & 0.43 & 0.33 & 0.31 & 0.33 & 0.26 & 0.22 & 0.32 & \underline{0.56} & \underline{0.54} \\
        \textbf{NW} & 0.19 & 0 & 0.18 & 0.21 & 0.26 & 0.33 & \textbf{0.36} & 0.28 & 0.30 & 0.23 & 0.29 & 0.27 & 0.15 & 0.26 & 0.19 & 0.35 & 0.32 \\
        \rowcolor[gray]{0.9}
        \textbf{OD} & \textbf{0.88} & 0 & 0.05 & 0.08 & 0.16 & 0.26 & 0.42 & 0.57 & 0.55 & 0.12 & 0.57 & 0.57 & 0.71 & 0.64 & 0.80 & 0.47 & 0.29 \\
        \textbf{OZ} & 0.39 & 0.33 & 0.31 & 0.28 & 0.25 & 0.26 & 0.30 & 0.33 & 0.33 & 0.33 & 0.32 & 0.33 & 0.34 & 0.36 & 0.34 & 0.25 & 0.21 \\
        \rowcolor[gray]{0.9}
        \textbf{PO} & -0.02 & 0 & 0.15 & 0.22 & 0.30 & 0.35 & \textbf{0.41} & 0.33 & 0.40 & 0.30 & 0.35 & 0.35 & 0.26 & 0.33 & 0.06 & 0.40 & 0.38 \\ 
        \textbf{PS} & -0.04 & -0.01 & 0.06 & 0.12 & \underline{0.16} & \underline{0.16} & 0.09 & 0.13 & 0.13 & 0.10 & 0.12 & 0.13 & 0.12 & 0.13 & 0.09 & 0.14 & \underline{0.16} \\
        \rowcolor[gray]{0.9}
        \textbf{RT} & -0.11 & 0 & 0.29 & 0.62 & 0.61 & 0.57 & 0.71 & 0.26 & 0.51 & 0.44 & 0.28 & 0.51 & 0.47 & 0.37 & 0.56 & \textbf{0.75} & 0.65 \\
        \textbf{SM} & -0.10 & -0.10 & 0.20 & 0.42 & 0.50 & 0.53 & 0.54 & 0.46 & 0.38 & 0.31 & 0.46 & 0.30 & 0.54 & 0.48 & \textbf{0.61} & 0.55 & 0.57 \\
        \rowcolor[gray]{0.9}
        \textbf{SL} & -0.01 & 0 & 0.61 & 0.68 & 0.73 & 0.76 & 0.82 & 0.56 & 0.71 & 0.67 & 0.59 & 0.50 & 0.66 & 0.64 & 0.67 & \textbf{0.84} & 0.80 \\
        \textbf{UW} & -0.02 & -0.02 & 0.22 & 0.34 & 0.40 & 0.43 & 0.35 & 0.21 & 0.31 & 0.10 & 0.21 & 0.34 & 0.45 & 0.47 & \textbf{0.49} & 0.41 & 0.47 \\
        \rowcolor[gray]{0.9}
        \textbf{YG} & -0.03 & 0 & 0.12 & 0.26 & 0.36 & 0.47 & 0.49 & 0.07 & 0.07 & 0.01 & 0.06 & 0.12 & 0.09 & 0.07 & 0.07 & \textbf{0.59} & 0.55 \\
        \midrule
        \textbf{Rank} & 13.67 & 15.71 & 10.85 & 8.5 & 6.93 & 5.21 & 7.02 & 11.56 & 9.19 & 10.31 & 9.81 & 9.42 & 8.69 & 8.22 & 7.90 & 5.21  & \textbf{4.79} \\ 
        \bottomrule
        \multicolumn{18}{l}{
            $^*$ Dataset not large enough to statistically distinguish performance over entire stream.
        }
    \end{tabular}
    \end{adjustbox}
\end{table}

\end{landscape}

\begin{landscape}
\begin{table}[htbp]
    \centering
    \caption{Ranks for each classifier (column) and data stream investigated (row). Ranks were determined using fractional ranking where methods that are not significantly different according to a Mann-Whitney U test ($p=0.05$) are given the same rank.}
    \label{tab:results_main_ranks}
    \begin{adjustbox}{angle=0}
    \begin{tabular}{l|c|c|c|c|c|c|c|c|c|c|c|c|c|c|c|c|c}
    \toprule
        \multicolumn{1}{c|}{} & \multicolumn{2}{c|}{\textbf{Baselines}}  & \multicolumn{5}{c|}{\textbf{Sliding Window}} & \multicolumn{3}{c|}{\textbf{Incremental}}  & \multicolumn{5}{c|}{\textbf{Drift Detectors}} & \multicolumn{2}{c}{\textbf{Ours}} \\
        \cmidrule{2-18}
        \multicolumn{1}{c|}{} & \textbf{LC} & \textbf{MC} & \textbf{DT$_{50}$} & \textbf{DT$_{100}$} & \textbf{DT$_{250}$} & \textbf{DT$_{500}$} & \textbf{DT$_{\infty}$} & \textbf{HT} & \textbf{EFDT} & \textbf{PLS} & \textbf{HAT} & \textbf{LAST} & \textbf{$\text{HT}_{\text{ADWIN}}$} & \textbf{$\text{HT}_{\text{DDM}}$}  & \textbf{$\text{HT}_{\text{D3}}$} & \textbf{TNT} & \textbf{TNT$_{500}$} \\
        \midrule
        \textbf{AF} & 16 & 17 & 14 & 10.5 & 4 & 4 & 4 & 10.5 & 10.5 & 10.5 & 10.5 & 4 & 4 & 10.5 & 15 & 4 & 4 \\
        \rowcolor[gray]{0.9}
        \textbf{CA} & 17 & 16 & 7 & 7 & 5 & 3.5 & 1.5 & 12 & 12 & 12 & 12 & 7 & 12 & 12 & 12 & 1.5 & 3.5 \\
        \textbf{CH} & 9 & 9 & 9 & 9 & 9 & 9 & 9 & 9 & 9 & 9 & 9 & 9 & 9 & 9 & 9 & 9 & 9 \\
        \rowcolor[gray]{0.9}
        \textbf{EL}  & 2 & 17 & 5 & 7.5 & 10.5 & 10.5 & 16 & 14 & 7.5 & 14 & 5 & 10.5 & 5 & 2 & 2 & 14 & 10.5 \\
        \textbf{FC} & 1 & 17 & 15 & 16 & 12 & 4.5 & 14 & 10 & 7.5 & 6 & 9 & 4.5 & 11 & 2 & 7.5 & 13 & 3 \\
        \rowcolor[gray]{0.9}
        \textbf{GS}  & 15.5 & 15.5 & 7.5 & 7.5 & 7.5 & 7.5 & 3 & 15.5 & 10.5 & 12.5 & 15.5 & 12.5 & 10.5 & 5 & 1 & 2 & 4 \\
        \textbf{HB} & 13 & 13 & 7 & 7 & 3 & 3 & 1 & 17 & 16 & 3 & 13 & 7 & 13 & 7 & 13 & 7 & 10 \\
        \rowcolor[gray]{0.9}
        \textbf{IA} & 16 & 17 & 11 & 5 & 3 & 2 & 13 & 15 & 12 & 8 & 8 & 14 & 4 & 10 & 8 & 6 & 1 \\
        \textbf{IG} & 16.5 & 16.5 & 11 & 5 & 3 & 2 & 11 & 11 & 11 & 11 & 11 & 11 & 11 & 11 & 6 & 4 & 1 \\
        \rowcolor[gray]{0.9}
        \textbf{II}  & 16 & 17 & 15 & 14 & 11 & 6 & 12 & 13 & 9.5 & 1.5 & 8 & 9.5 & 1.5 & 6 & 3 & 6 & 4 \\
        \textbf{IR} & 16 & 17 & 6 & 4 & 3 & 2 & 15 & 14 & 12 & 10 & 10 & 13 & 5 & 10 & 7.5 & 7.5 & 1 \\
        \rowcolor[gray]{0.9}
        \textbf{KS} & 16.5 & 16.5 & 8 & 6.5 & 4 & 4 & 4 & 12.5 & 9.5 & 15 & 12.5 & 12.5 & 12.5 & 9.5 & 6.5 & 2 & 1 \\
        \textbf{LX} & 16.5 & 16.5 & 8 & 8 & 8 & 8 & 8 & 8 & 8 & 8 & 8 & 8 & 8 & 8 & 8 & 8 & 8 \\
        \rowcolor[gray]{0.9}
        \textbf{MS} & 16.5 & 16.5 & 7.5 & 3.5 & 3.5 & 3.5 & 3.5 & 14.5 & 7.5 & 10.5 & 10.5 & 10.5 & 13 & 14.5 & 10.5 & 3.5 & 3.5 \\
        \textbf{NW} & 14 & 17 & 14 & 11.5 & 8.5 & 3 & 1 & 8.5 & 5 & 11.5 & 5 & 8.5 & 16 & 8.5 & 14 & 2 & 5 \\
        \rowcolor[gray]{0.9}
        \textbf{OD} & 1 & 17 & 16 & 15 & 13.5 & 12 & 10 & 6 & 8 & 13.5 & 6 & 6 & 3 & 4 & 2 & 9 & 11 \\
        \textbf{OZ} & 8 & 8 & 8 & 8 & 16.5 & 8 & 8 & 8 & 8 & 8 & 8 & 8 & 8 & 8 & 8 & 8 & 16.5 \\
        \rowcolor[gray]{0.9}
        \textbf{PO} & 17 & 16 & 14 & 13 & 11 & 6 & 1 & 8 & 3 & 10 & 6 & 6 & 12 & 9 & 15 & 2 & 4 \\
        \textbf{PS} & 17 & 16 & 15 & 7.5 & 2 & 2 & 12.5 & 7.5 & 7.5 & 12.5 & 7.5 & 7.5 & 11 & 7.5 & 14 & 4 & 2 \\
        \rowcolor[gray]{0.9}
        \textbf{RT} & 17 & 16 & 13 & 4 & 5 & 6 & 2 & 15 & 8.5 & 11 & 14 & 8.5 & 10 & 12 & 7 & 1 & 3 \\
        \textbf{SM} & 17 & 16 & 15 & 11 & 7 & 6 & 4 & 9.5 & 12 & 13 & 9.5 & 14 & 5 & 8 & 1 & 3 & 2 \\
        \rowcolor[gray]{0.9}
        \textbf{SL} & 16.5 & 16.5 & 12 & 8 & 5 & 4 & 2 & 14.5 & 6 & 8 & 13 & 14.5 & 10.5 & 10.5 & 8 & 1 & 3 \\
        \textbf{UW} & 16.5 & 16.5 & 12 & 9.5 & 6.5 & 5 & 9.5 & 14 & 9.5 & 14 & 14 & 9.5 & 3 & 3 & 1 & 6.5 & 3 \\
        \rowcolor[gray]{0.9}
        \textbf{YG} & 16.5 & 16.5 & 10.5 & 6 & 5 & 3.5 & 3.5 & 10.5 & 10.5 & 15 & 10.5 & 10.5 & 10.5 & 10.5 & 10.5 & 1 & 2 \\     
        \bottomrule
    \end{tabular}
    \end{adjustbox}
\end{table}

\end{landscape}


\subsection{Postmortem} \label{sect:app_postmortem}

In this final Section, we provide a \textit{postmortem} of the research conducted in this work. It is somewhat informal in nature, and contains insights that we either did not have time to validate, or did not include in the main paper because they were not, in our opinion, significant enough. This Section proceeds as follows: Section \ref{sect:app_postmortem_param_tune} discusses the consequences of the parameter tuning process in this work, Section \ref{sect:app_postmortem_incremental_learners} discusses some of the nuances regarding the deployment of Incremental Stream Learners (specifically Decision Trees) and Section \ref{sect:app_postmortem_TNT} reflects on Time Neutralising Trees (TNT) and provides insight into potential future research directions. 

\subsubsection{Synthetic Streams Parameter Tuning and Active Forgetting} \label{sect:app_postmortem_param_tune}

In this work, we used synthetic data streams to parameter tune TNT to avoid data leakage and bias which are known problems in Stream Learning research \cite{SouzaReisMaletzke2020}. The streams we used are discussed in Section \ref{sect:app_param_tune}. We note that using synthetic streams has received criticism recently \cite{SouzaReisMaletzke2020} for being unrealistic, which this section reflects on. Based on the results of parameter tuning TNT using the synthetic streams, we settled on a series of hyper-parameters that were "desirable". However, it is worth considering if those values were "desirable" on the USP DS repository. Our intuition tells use that $\lambda = 0.2$ is likely too conservative for streams where active forgetting is necessary. This is quite clear on the \verb|Hyperplane| stream (Fig. \ref{fig:app_hyperplane_sa}) where the optimal $\lambda$ was 0.0. That begs the question, when is active forgetting useful? An informal investigation reveals that on many of the data streams we evaluated, active forgetting was only contextually, and typically transiently useful. For example, immediately after an abrupt drift, active forgetting is useful, but, as the stream progressed, it was less useful as TNT (and the other Batch DTs) were simply able to separate the change in distributions. This is evident by the high mean ranking of DT$_{\infty}$ which, to us, implies that simply training on all of the data is a very viable strategy in many stream settings. More specifically, that may imply that the "concept drift" in many streams does not result in conflicting feature subspaces along time (i.e., where active forgetting is useful). 

Conversely, it is clear from our results that passive forgetting (i.e., sliding windows) is useful. This is shown by the fact that TNT$_{500}$, DT$_{500}$, and DT$_{250}$ ranked 1st, (tied) 2nd, and 4th resp. It is also a core assumption of data stream learning that recent data is more relevant to future predictions than past data. That begs another question, why? Or more specifically, why was the time filtering active forgetting of TNT not enough? We have already mentioned that $\lambda$ may have been tuned too conservatively, but another theory is that near-optimal active forgetting is not achievable through observations of the training instances alone. In other words, simply looking at historical instances of a stream is not enough to determine which of those instances is useful to train on. Instead, understanding how the streams drifts / evolves over time would solve this problem. Any further speculation is beyond the scope of this paper, and is fundamentally antithetical to the reactive nature of Stream Learning. In fact, it is more inline with the proactive approaches proposed in Drift Modelling. Nevertheless, we believe this line of thought may be useful in future work.

\subsubsection{Incremental Stream Learners} \label{sect:app_postmortem_incremental_learners}

This Section reflects on the use of traditional Incremental Stream Learners (specifically Hoeffding Trees and their variants) (HTs) in the modern age. Our results, combined with the results of Gower-Winter et al. \cite{GowerGroenKrempl2026}, paint a very convincing picture that incremental decisions trees are insufficient in the modern era where both compute and memory is more readily available. We are not saying that these methods are useless, but rather that their usage should be reserved for the settings upon which they were developed (e.g., low resource Edge Computing). Futhermore, we also want to take note of the settings where HTs perfromed well. These were the \verb|Electricity|, \verb|Forest Cover Type| and \verb|Outdoor Objects| streams. Interestingly, these also happen to be the streams with the highest label autocorrelation (evident by the Kappa score of the Last Class (LC) classifier). Furthermore, \verb|Electricity|, \verb|Forest Cover Type| and \verb|Outdoor Objects| also happen to be some of the oldest data streams, and we can't help but think that data leakage (or at least some bias) is present in the design of these HT architectures. It is also worth noting: In this work, HT, and its variants, train immediately after they make a prediction on an instance. This is obviously unrealistic in many settings as labels do not arrive instantly, but it is standard practice. It is almost certain that if we restricted the training periods of the incremental DTs to follow that of the Batch DTs (training only every 50 instances), their performance would drop, further exacerbating the divide between the Batch and Stream learners \cite{GowerGroenKrempl2026}.

To prove this point, we ran additional experiments on the aforementioned streams: \verb|Electricity| (EL), \verb|Forest Cover Type| (FC) and \verb|Outdoor Objects| (OD) using the best supervised detector-based stream learning approach from our main evaluation: DDM. In these experiments, DDM was only able to update every 50 instances training on the last 50 instances in the stream, thus matching the updating cadence of the batch DTs. Table \ref{tab:app_verf_latency} reports the results which clearly shows the benefit of updating after every instance\footnote{These are the non-posthoc adjusted Kappa Scores caused by the river bug which affected the main results in Table \ref{tab:results_main}.}. In fact, the performance drop is so extreme in all cases, that DDM would have ranked worse than most of the batch learning methods on all three streams evaluated. These results further support our claim that incremental Decision Trees are simply inferior to their Batch Learning counterparts. While we cannot say for certain, this performance drop is likely for all other stream learning methods investigated \cite{GowerGroenKrempl2026}.

\begin{table}[ht]
    \centering
    \caption{Average Kappa Score of Hoeffding Tree (with DDM) trained incrementally (test-then-train) and trained in a batch fashion (which simulates label delay, and mimics the training regime of the batch Decision Trees investigated in this work). Results show clear the benefit of training immediately after making a prediction on an instance (i.e., with no label delay).}
    \label{tab:app_verf_latency}
    \begin{tabular}{l|c|c|c}
    \toprule
        \multicolumn{1}{c|}{} & \textbf{EL} & \textbf{FC} & \textbf{OD}  \\
        \midrule
        \rowcolor[gray]{0.9}
        \textbf{DDM} & 0.61 & 0.42 & 0.63 \\
        \textbf{DDM (Batched)} & 0.45 & 0.07 & 0.19 \\
        \bottomrule
    \end{tabular}
\end{table}

\subsubsection{Time Neutralising Trees (TNT)} \label{sect:app_postmortem_TNT}

This Section reflects on our implementation of TNT. Overall, we are happy with the results of TNT on the USP DS repository. Both variants of TNT ranked well overall: 1st for TNT$_{500}$ and 2nd (tied) for the base variant (TNT). We believe our approach stands as the new best-in-class for single Decision Trees in stream learning settings with concept drift. That being said, we must acknowledge that the current state-of-the-art stream learners are all tree ensemble approaches such as ARF \cite{GomesBifetRead2017}, AMF \cite{MourtadaGaiffasScornet2021}, SRP \cite{GomesReadBifet2019} and SGBT \cite{GunasekaraPfahringerGomes2024}. This presents us with a natural next step in the development of concept drift robust batch learners (i.e., An ensemble approach for TNT). We did explore this idea initially. However, ensemble stream learning approaches (e.g., ARF) adopt a dynamic approach in the way the partition the feature space for each base tree. More specifically, there is kind of evolutionary survival of the fittest game being played implicitly whereby poorly performing base trees are replaced until a stable variant emerges. Achieving a batch learning equivalent of this dynamic is far beyond the scope of this paper, but is something that we will explicitly tackle in the future.

Lastly, we do want to acknowledge that the current implementation of TNT has a potential fail case. This has to do with the way TNT merges leaf nodes during the filtering process. In this version of TNT, we perform a Mann-Whitney U test on the label distributions of each leaf node using the right most leaf node (i.e., the most recent samples in the stream) as reference. If they are sufficiently similar, then the nodes are "merged" (i.e., instances from the older leaf node are not filtered out). This approach worked well in this work, but in a setting where a subspace being evaluated for filtering consists of instances with the same label distribution P(Y), but different posterior distribution P(Y|X), leaf nodes will still be merged even if they shouldn't be. A simple example of this is a binary "class flip" whereby, for some feature subspace: P(X) and P(Y) remain the same, but P(Y|X) flips meaning that instances that were "positive" are now "negative" and vice-versa. A more robust solution to this limitation would be to train a lightweight model on the instances in the right most leaf node of the filter subtree, and only merge other leaf nodes if the model achieves a metric score > some threshold. This would, however, come with additional compute cost. Nevertheless, we reiterate that this version of TNT is capable of achieving state-of-art results.

\newpage
\input{checklist.tex}

\end{document}

%% file: checklist.tex
\section*{NeurIPS Paper Checklist}

\begin{enumerate}

\item {\bf Claims}
    \item[] Question: Do the main claims made in the abstract and introduction accurately reflect the paper's contributions and scope?
    \item[] Answer: \answerYes{} 
    \item[] Justification: Our main research goals were: (1) Stream learning should move towards batch style training regimes and (2) our proposed method TNT is a state-of-the-art stream classification method. With our extensive experimentation, we believe the contents of the paper accurately reflect the claims made in the abstract and introduction.

\item {\bf Limitations}
    \item[] Question: Does the paper discuss the limitations of the work performed by the authors?
    \item[] Answer: \answerYes{} 
    \item[] Justification: We are quite honest that the parameter tuning process resulted in hyper-parameters that were likely too conservative. We acknowledge a potential failure case (Section \ref{sect:app_postmortem_TNT}) for TNT, and repeatedly state that our results are only relevant in settings where storing instances for a period of time is viable. Moreover, we also clearly state that TNT is a state-of-the-art single decision method, and that more work needs to be done in order to transfer the time filtering process to ensemble architectures.

\item {\bf Theory assumptions and proofs}
    \item[] Question: For each theoretical result, does the paper provide the full set of assumptions and a complete (and correct) proof?
    \item[] Answer: \answerYes{} 
    \item[] Justification: From our perspective, Section \ref{sect:method_motive} provides theoretical results, but they are largely based on extensive theoretical work already done in Information Theory. We simply cite the relevant sources. We do also include empirical evidence that the theoretical results are in fact valid (Section \ref{sect:app_MI_EMP}). 

    \item {\bf Experimental result reproducibility}
    \item[] Question: Does the paper fully disclose all the information needed to reproduce the main experimental results of the paper to the extent that it affects the main claims and/or conclusions of the paper (regardless of whether the code and data are provided or not)?
    \item[] Answer: \answerYes{} 
    \item[] Justification: Besides the fact that we have included the source code, Sections \ref{sect:exp_design} and \ref{sect:app_exp_details} cover, in great detail, the information needed to reproduce this work. Section \ref{sect:method_TNT} also describes our method and we include pseudocode for the relevant functions.

\item {\bf Open access to data and code}
    \item[] Question: Does the paper provide open access to the data and code, with sufficient instructions to faithfully reproduce the main experimental results, as described in supplemental material?
    \item[] Answer: \answerYes{} 
    \item[] Justification: We have included the source code as supplementary material, and cite all relevant datasets.

\item {\bf Experimental setting/details}
    \item[] Question: Does the paper specify all the training and test details (e.g., data splits, hyperparameters, how they were chosen, type of optimizer) necessary to understand the results?
    \item[] Answer: \answerYes{} 
    \item[] Justification: Not only do we state the hyper-parameters selected (Section \ref{sect:exp_design_param_tune}), we also include the range of parameters explored for TNT (Section \ref{sect:app_param_tune}), and state the hyper-parameters for all other methods used in our evaluation (Section \ref{sect:app_exp_details}).

\item {\bf Experiment statistical significance}
    \item[] Question: Does the paper report error bars suitably and correctly defined or other appropriate information about the statistical significance of the experiments?
    \item[] Answer: \answerYes{} 
    \item[] Justification: We performed statistical tests for comparing all methods in our evaluation (Section \ref{sect:exp_design}). Error bars are included but also context dependent. For example, our main results (Table \ref{tab:results_main}) do not include standard deviations because it would make the table unreadable, so we instead show significance through bolding the statistically best method(s). Other results include error bars where the add to the interpretability of said results (e.g., Fig. \ref{fig:app_mi_collapse}).

\item {\bf Experiments compute resources}
    \item[] Question: For each experiment, does the paper provide sufficient information on the computer resources (type of compute workers, memory, time of execution) needed to reproduce the experiments?
    \item[] Answer: \answerYes{} 
    \item[] Justification: Section \ref{sect:app_exp_details} lists the compute resources used. We do provide the compute times for TNT in the form of a performance analysis, but we did not keep track of the overall runtime of the entire experimental process. Primarily due to the fact that most experiments were run on a laptop that was also being used for day-to-day tasks, resulting in fluctuating compute resources that would have made any compute time reporting inconsistent and unreflective of the compute times had they been done on a dedicated machine.
    
\item {\bf Code of ethics}
    \item[] Question: Does the research conducted in the paper conform, in every respect, with the NeurIPS Code of Ethics \url{https://neurips.cc/public/EthicsGuidelines}?
    \item[] Answer: \answerYes{} 
    \item[] Justification: Justification for this section is unclear because it is entirely personal opinion. We read the Code of Ethics, we believe we have conformed to them.

\item {\bf Broader impacts}
    \item[] Question: Does the paper discuss both potential positive societal impacts and negative societal impacts of the work performed?
    \item[] Answer: \answerNA{} 
    \item[] Justification: It can be argued that every new Machine Learning algorithm carries with it the potential to do harm. Beyond this, TNT does not explicitly provide any positive and/or negative societal impacts. 
    
\item {\bf Safeguards}
    \item[] Question: Does the paper describe safeguards that have been put in place for responsible release of data or models that have a high risk for misuse (e.g., pre-trained language models, image generators, or scraped datasets)?
    \item[] Answer: \answerNA{} 
    \item[] Justification: See previous question. We also note that all real-world data streams investigated come from a repository that is publicly accessible to everyone \cite{SouzaReisMaletzke2020}.

\item {\bf Licenses for existing assets}
    \item[] Question: Are the creators or original owners of assets (e.g., code, data, models), used in the paper, properly credited and are the license and terms of use explicitly mentioned and properly respected?
    \item[] Answer: \answerYes{} 
    \item[] Justification: All relevant software packages, algorithms and datasets are explicitly cited and acknowledged (Section \ref{sect:app_exp_details}). Any version relevant dependencies can be seen in provided source code.

\item {\bf New assets}
    \item[] Question: Are new assets introduced in the paper well documented and is the documentation provided alongside the assets?
    \item[] Answer: \answerYes{} 
    \item[] Justification: Code comments and usage instructions for TNT are provided with the source code.

\item {\bf Crowdsourcing and research with human subjects}
    \item[] Question: For crowdsourcing experiments and research with human subjects, does the paper include the full text of instructions given to participants and screenshots, if applicable, as well as details about compensation (if any)? 
    \item[] Answer: \answerNA{} 
    \item[] Justification: No human subjects in this work.

\item {\bf Institutional review board (IRB) approvals or equivalent for research with human subjects}
    \item[] Question: Does the paper describe potential risks incurred by study participants, whether such risks were disclosed to the subjects, and whether Institutional Review Board (IRB) approvals (or an equivalent approval/review based on the requirements of your country or institution) were obtained?
    \item[] Answer: \answerNA{} 
    \item[] Justification: See previous question.

\item {\bf Declaration of LLM usage}
    \item[] Question: Does the paper describe the usage of LLMs if it is an important, original, or non-standard component of the core methods in this research? Note that if the LLM is used only for writing, editing, or formatting purposes and does \emph{not} impact the core methodology, scientific rigor, or originality of the research, declaration is not required.
    \item[] Answer: \answerNA{} 
    \item[] Justification: We wish to be transparent that Gemini 3.1 Pro was used to create style sheets for the plots included in this work (i.e., Matplotlib configurations / setups that looked research paper-quality), to optimise boiler plate code that was slowing down evaluations and to convert some sketch equations into  \LaTeX  compliant equations. There was also implicit use of code autocomplete provided by the IDE used (PyCharm). Any and all output produced by LLMs was reviewed by a human. No data, core functional code, or any written text in this paper was generated using LLMs

\end{enumerate}

%% file: references.bib
@inproceedings{AssisBarddalEnembreck2024,
  title={Just change on change: Adaptive splitting time for decision trees in data stream classification},
  author={Assis, Daniel Nowak and Barddal, Jean Paul and Enembreck, Fabr{\'\i}cio},
  booktitle={Proceedings of the 39th ACM/SIGAPP Symposium on Applied Computing},
  pages={351--357},
  year={2024}
}

@article{BenavoliCoraniMangili2016,
  title={Should we really use post-hoc tests based on mean-ranks?},
  author={Benavoli, Alessio and Corani, Giorgio and Mangili, Francesca},
  journal={The Journal of Machine Learning Research},
  volume={17},
  number={1},
  pages={152--161},
  year={2016}
}

@inproceedings{BifetGavalda2007,
  title={Learning from time-changing data with adaptive windowing},
  author={Bifet, Albert and Gavalda, Ricard},
  booktitle={Proc. of the 2007 {SIAM} Int. Conf. on Data Mining},
  pages={443--448},
  year={2007},
  organization={SIAM}
}

@inproceedings{BifetGalvalda2009,
  title={Adaptive learning from evolving data streams},
  author={Bifet, Albert and Gavalda, Ricard},
  booktitle={Int. Symposium on Intelligent Data Analysis},
  pages={249--260},
  year={2009},
  organization={Springer}
}

@article{BlackardDean1999,
  title={Comparative accuracies of artificial neural networks and discriminant analysis in predicting forest cover types from cartographic variables},
  author={Blackard, Jock A and Dean, Denis J},
  journal={Computers and Electronics in Agriculture},
  volume={24},
  number={3},
  pages={131--151},
  year={1999},
  publisher={Elsevier}
}

@article{Breiman2001,
  title={Random forests},
  author={Breiman, Leo},
  journal={Machine learning},
  volume={45},
  number={1},
  pages={5--32},
  year={2001},
  publisher={Springer}
}

@article{CerqueiraGomesBifet2023,
  title={{STUDD}: A student--teacher method for unsupervised concept drift detection},
  author={Cerqueira, Vitor and Gomes, Heitor Murilo and Bifet, Albert and Torgo, Luis},
  journal={Machine Learning},
  volume={112},
  number={11},
  pages={4351--4378},
  year={2023},
  publisher={Springer}
}

@article{Cohen1960,
  title={A coefficient of agreement for nominal scales},
  author={Cohen, Jacob},
  journal={Educational and psychological measurement},
  volume={20},
  number={1},
  pages={37--46},
  year={1960},
  publisher={Sage Publications Sage CA: Thousand Oaks, CA}
}

@article{CoverThomas2006,
  title={Elements of Information Theory},
  author={Cover, Thomas M and Thomas, Joy A},
  year={2006}
}

@misc{DauKeoghKamgar2018,
    title = {The {UCR} {T}ime {S}eries {C}lassification {A}rchive},
    author = {Dau, Hoang Anh and Keogh, Eamonn and Kamgar, Kaveh and Yeh, Chin-Chia Michael and Zhu, Yan 
                  and Gharghabi, Shaghayegh and Ratanamahatana, Chotirat Ann and Yanping and Hu, Bing 
                  and Begum, Nurjahan and Bagnall, Anthony and Mueen, Abdullah and Batista, Gustavo and Hexagon-ML},
    year = {2018},
}

@article{Demsar2006,
  title={Statistical comparisons of classifiers over multiple data sets},
  author={Dem{\v{s}}ar, Janez},
  journal={The Journal of Machine learning research},
  volume={7},
  number={1},
  pages={1--30},
  year={2006}
}

@article{DheeruTaniskidou2017,
  title={{UCI} {M}achine {L}earning {R}epository},
  author={Dheeru, Dua and Taniskidou, E Karra},
  year={2017},
  publisher={Irvine, CA, USA}
}

@article{DitzlerPolikar2012,
  title={Incremental learning of concept drift from streaming imbalanced data},
  author={Ditzler, Gregory and Polikar, Robi},
  journal={{IEEE} Transactions on Knowledge and Data Engineering},
  volume={25},
  number={10},
  pages={2283--2301},
  year={2012},
  publisher={IEEE}
}

@article{Friedman2001,
  title={Greedy function approximation: a gradient boosting machine},
  author={Friedman, Jerome H},
  journal={Annals of statistics},
  pages={1189--1232},
  year={2001},
  publisher={JSTOR}
}

@article{Friedman2002,
  title={Stochastic gradient boosting},
  author={Friedman, Jerome H},
  journal={Computational statistics \& data analysis},
  volume={38},
  number={4},
  pages={367--378},
  year={2002},
  publisher={Elsevier}
}

@inproceedings{GamaMedasCastillo2004,
  title={Learning with drift detection},
  author={Gama, Joao and Medas, Pedro and Castillo, Gladys and Rodrigues, Pedro},
  booktitle={Brazilian Symposium on Artificial Intelligence},
  pages={286--295},
  year={2004},
  organization={Springer}
}

@article{GamaZliobaiteBifet2014,
  title={A survey on concept drift adaptation},
  author={Gama, Jo{\~a}o and {\v{Z}}liobait{\.e}, Indr{\.e} and Bifet, Albert and Pechenizkiy, Mykola and Bouchachia, Abdelhamid},
  journal={{ACM} Computing Surveys ({CSUR})},
  volume={46},
  number={4},
  pages={1--37},
  year={2014},
  publisher={ACM New York, NY, USA}
}

@article{GomesBifetRead2017,
  title={Adaptive random forests for evolving data stream classification},
  author={Gomes, Heitor M and Bifet, Albert and Read, Jesse and Barddal, Jean Paul and Enembreck, Fabr{\'\i}cio and Pfharinger, Bernhard and Holmes, Geoff and Abdessalem, Talel},
  journal={Machine Learning},
  volume={106},
  number={9},
  pages={1469--1495},
  year={2017},
  publisher={Springer}
}

@article{GomesBarddalEnembreck2017,
  title={A survey on ensemble learning for data stream classification},
  author={Gomes, Heitor Murilo and Barddal, Jean Paul and Enembreck, Fabr{\'\i}cio and Bifet, Albert},
  journal={ACM Computing Surveys (CSUR)},
  volume={50},
  number={2},
  pages={1--36},
  year={2017},
  publisher={ACM New York, NY, USA}
}

@inproceedings{GomesReadBifet2019,
  title={Streaming random patches for evolving data stream classification},
  author={Gomes, Heitor Murilo and Read, Jesse and Bifet, Albert},
  booktitle={2019 IEEE international conference on data mining (ICDM)},
  pages={240--249},
  year={2019},
  organization={IEEE}
}

@inproceedings{GowerGroenKrempl2026,
  title={The Window Dilemma: Why Concept Drift Detection is Ill-Posed},
  author={Gower-Winter, Brandon and Groen, Misja and Krempl, Georg},
  booktitle={International Symposium on Intelligent Data Analysis},
  pages={371--384},
  year={2026},
  organization={Springer}
}

@inproceedings{GozuaccikBuyukccakirBonab2019,
  title={Unsupervised concept drift detection with a discriminative classifier},
  author={G{\"o}z{\"u}a{\c{c}}{\i}k, {\"O}mer and B{\"u}y{\"u}k{\c{c}}ak{\i}r, Alican and Bonab, Hamed and Can, Fazli},
  booktitle={Proc. of the 28th {ACM} Int. Conf. on Information and Knowledge Management},
  pages={2365--2368},
  year={2019}
}

@article{GunasekaraPfahringerGomes2024,
  title={Gradient boosted trees for evolving data streams},
  author={Gunasekara, Nuwan and Pfahringer, Bernhard and Gomes, Heitor and Bifet, Albert},
  journal={Machine Learning},
  volume={113},
  number={5},
  pages={3325--3352},
  year={2024},
  publisher={Springer}
}

@article{Harries1999,
  title={Splice-2 comparative evaluation: Electricity pricing},
  author={Harries, Michael},
  year={1999},
  publisher={University of New South Wales, School of Computer Science and Engineering}
}

@inproceedings{HeydenGomesFouche2024,
  title={Leveraging plasticity in incremental decision trees},
  author={Heyden, Marco and Gomes, Heitor Murilo and Fouch{\'e}, Edouard and Pfahringer, Bernhard and B{\"o}hm, Klemens},
  booktitle={Joint European conference on machine learning and knowledge discovery in databases},
  pages={38--54},
  year={2024},
  organization={Springer}
}

@inproceedings{HinderVaquetBrinkrolf2023,
  title={On the change of decision boundary and loss in learning with concept drift},
  author={Hinder, Fabian and Vaquet, Valerie and Brinkrolf, Johannes and Hammer, Barbara},
  booktitle={Int. Symposium on Intelligent Data Analysis},
  pages={182--194},
  year={2023},
  organization={Springer}
}

@article{HinderVaquetHammer2024,
  title={One or two things we know about concept drift—a survey on monitoring in evolving environments. Part A: detecting concept drift},
  author={Hinder, Fabian and Vaquet, Valerie and Hammer, Barbara},
  journal={Frontiers in Artificial Intelligence},
  volume={7},
  pages={1330257},
  year={2024},
  publisher={Frontiers Media SA}
}

@inproceedings{HultenSpencerDomingos2001,
  title={Mining time-changing data streams},
  author={Hulten, Geoff and Spencer, Laurie and Domingos, Pedro},
  booktitle={Proc. of the seventh ACM SIGKDD Int. Conf. on Knowledge Discovery and Data Mining},
  pages={97--106},
  year={2001}
}

@article{KhamassiSayedHammami2018,
  title={Discussion and review on evolving data streams and concept drift adapting},
  author={Khamassi, Imen and Sayed-Mouchaweh, Moamar and Hammami, Moez and Gh{\'e}dira, Khaled},
  journal={Evolving systems},
  volume={9},
  number={1},
  pages={1--23},
  year={2018},
  publisher={Springer}
}

@article{KremplZliobaiteBrzezinski2014,
  title={Open challenges for data stream mining research},
  author={Krempl, Georg and {\v{Z}}liobaite, Indre and Brzezi{\'n}ski, Dariusz and H{\"u}llermeier, Eyke and Last, Mark and Lemaire, Vincent and Noack, Tino and Shaker, Ammar and Sievi, Sonja and Spiliopoulou, Myra and others},
  journal={ACM SIGKDD explorations newsletter},
  volume={16},
  number={1},
  pages={1--10},
  year={2014},
  publisher={ACM New York, NY, USA}
}

@inproceedings{Koychev2002,
  title={Tracking changing user interests through prior-learning of context},
  author={Koychev, Ivan},
  booktitle={International Conference on Adaptive Hypermedia and Adaptive Web-Based Systems},
  pages={223--232},
  year={2002},
  organization={Springer}
}

@article{KunchevaZliobaite2009,
  title={On the window size for classification in changing environments},
  author={Kuncheva, Ludmila I and {\v{Z}}liobait{\.e}, Indr{\.e}},
  journal={Intelligent Data Analysis},
  volume={13},
  number={6},
  pages={861--872},
  year={2009},
  publisher={SAGE Publications}
}

@article{LiuLuLiu2018,
  title={Accumulating regional density dissimilarity for concept drift detection in data streams},
  author={Liu, Anjin and Lu, Jie and Liu, Feng and Zhang, Guangquan},
  journal={Pattern Recognition},
  volume={76},
  pages={256--272},
  year={2018},
  publisher={Elsevier}
}

@inproceedings{LosingHammerWersing2015,
  title={Interactive online learning for obstacle classification on a mobile robot},
  author={Losing, Viktor and Hammer, Barbara and Wersing, Heiko},
  booktitle={2015 International Joint Conference on Neural Networks (IJCNN)},
  pages={1--8},
  year={2015},
  organization={IEEE}
}

@inproceedings{LosingHammerWersing2016,
  title={KNN classifier with self adjusting memory for heterogeneous concept drift},
  author={Losing, Viktor and Hammer, Barbara and Wersing, Heiko},
  booktitle={IEEE 16th Int. Conf. on Data Mining (ICDM)},
  pages={291-300},
  year={2016},
  organization={IEEE}
}

@article{LuLiuDong2018,
  title={Learning under concept drift: A review},
  author={Lu, Jie and Liu, Anjin and Dong, Fan and Gu, Feng and Gama, Joao and Zhang, Guangquan},
  journal={IEEE Transactions on Knowledge and Data Engineering},
  volume={31},
  number={12},
  pages={2346--2363},
  year={2018},
  publisher={IEEE}
}

@inproceedings{LuLuLiu2025,
  title={Early concept drift detection via prediction uncertainty},
  author={Lu, Pengqian and Lu, Jie and Liu, Anjin and Zhang, Guangquan},
  booktitle={Proc. of the AAAI Conf. on Artificial Intelligence},
  volume={39},
  number={18},
  pages={19124--19132},
  year={2025}
}

@article{LukatsZielinskiHahn2025,
  title={A benchmark and survey of fully unsupervised concept drift detectors on real-world data streams},
  author={Lukats, Daniel and Zielinski, Oliver and Hahn, Axel and Stahl, Frederic},
  journal={Int. Journal of Data Science and Analytics},
  volume={19},
  number={1},
  pages={1--31},
  year={2025},
  publisher={Springer}
}

@inproceedings{ManapragadaWebbSalehi2018,
  title={Extremely fast decision tree},
  author={Manapragada, Chaitanya and Webb, Geoffrey I and Salehi, Mahsa},
  booktitle={Proceedings of the 24th ACM SIGKDD international conference on knowledge discovery \& data mining},
  pages={1953--1962},
  year={2018}
}

@article{MourtadaGaiffasScornet2021,
  title={AMF: Aggregated Mondrian forests for online learning},
  author={Mourtada, Jaouad and Ga{\"\i}ffas, St{\'e}phane and Scornet, Erwan},
  journal={Journal of the Royal Statistical Society Series B: Statistical Methodology},
  volume={83},
  number={3},
  pages={505--533},
  year={2021},
  publisher={Oxford University Press}
}

@article{Montiel2021,
  title={River: machine learning for streaming data in Python},
  author={Montiel, Jacob and Halford, Max and Mastelini, Saulo Martiello
          and Bolmier, Geoffrey and Sourty, Raphael and Vaysse, Robin and Zouitine, Adil
          and Gomes, Heitor Murilo and Read, Jesse and Abdessalem, Talel and others},
  year={2021}
}

@book{Niederreiter1992,
  title={Random number generation and quasi-Monte Carlo methods},
  author={Niederreiter, Harald},
  year={1992},
  publisher={SIAM}
}

@article{PedgregosaVaroquauxGramfort2011,
  title={Scikit-learn: Machine Learning in {P}ython},
  author={Pedregosa, F. and Varoquaux, G. and Gramfort, A. and Michel, V.
          and Thirion, B. and Grisel, O. and Blondel, M. and Prettenhofer, P.
          and Weiss, R. and Dubourg, V. and Vanderplas, J. and Passos, A. and
          Cournapeau, D. and Brucher, M. and Perrot, M. and Duchesnay, E.},
  journal={Journal of Machine Learning Research},
  volume={12},
  pages={2825--2830},
  year={2011}
}

@article{ReadZliobaite2025,
  title={Supervised Learning from Data Streams: An Overview and Update},
  author={Read, Jesse and {\v{Z}}liobait{\.e}, Indr{\.e}},
  journal={ACM Computing Surveys},
  year={2025},
  publisher={ACM New York, NY}
}

@inproceedings{ReisFlachMatwin2016fast,
  title={Fast unsupervised online drift detection using incremental kolmogorov-smirnov test},
  author={Reis, Denis Moreira Dos and Flach, Peter and Matwin, Stan and Batista, Gustavo},
  booktitle={Proceedings of the 22nd ACM SIGKDD international conference on knowledge discovery and data mining},
  pages={1545--1554},
  year={2016}
}

@article{SchlimmerGranger1986,
  title={Incremental learning from noisy data},
  author={Schlimmer, Jeffrey C and Granger Jr, Richard H},
  journal={Machine learning},
  volume={1},
  number={3},
  pages={317--354},
  year={1986},
  publisher={Springer}
}

@inproceedings{SouzaSilvaGamaBatista2015,
author = {Souza, Vin{\'\i}cius M. A. and Silva, Diego F. and Gama, Jo{\~a}o and Batista, Gustavo E. A. P. A.},
year = {2015},
pages = {873-881},
title = {Data Stream Classification Guided by Clustering on Nonstationary Environments and Extreme Verification Latency}
}

@article{Souza2018,
  title={Asphalt pavement classification using smartphone accelerometer and complexity invariant distance},
  author={Souza, Vinicius MA},
  journal={Engineering Applications of Artificial Intelligence},
  volume={74},
  pages={198--211},
  year={2018},
  publisher={Elsevier}
}

@article{SouzaReisMaletzke2020,
     title={Challenges in Benchmarking Stream Learning Algorithms with Real-world Data},
     author={Souza, Vinicius M. A. and Reis, Denis M. dos and Maletzke, Andr{\'e} G. and Batista, Gustavo E. A. P. A.},
     journal={Data Mining and Knowledge Discovery},
     pages={1805-1858},
     volume={34},
     year={2020}
}

@article{SouzaParmezanChowdhury2021,
  title={Efficient unsupervised drift detector for fast and high-dimensional data streams},
  author={Souza, Vinicius MA and Parmezan, Antonio RS and Chowdhury, Farhan A and Mueen, Abdullah},
  journal={Knowledge and Information Systems},
  volume={63},
  number={6},
  pages={1497--1527},
  year={2021},
  publisher={Springer}
}

@inproceedings{StreetKim2001,
  title={A streaming ensemble algorithm (SEA) for large-scale classification},
  author={Street, W Nick and Kim, YongSeog},
  booktitle={Proceedings of the seventh ACM SIGKDD international conference on Knowledge discovery and data mining},
  pages={377--382},
  year={2001}
}

@inproceedings{UlanovaBegumShokoohi2016,
  title={Clustering in the face of fast changing streams},
  author={Ulanova, Liudmila and Begum, Nurjahan and Shokoohi-Yekta, Mohammad and Keogh, Eamonn},
  booktitle={Proceedings of the 2016 SIAM international conference on data mining},
  pages={1--9},
  year={2016},
  organization={SIAM}
}

@article{VergaraVembuAyhan2012,
  title={Chemical gas sensor drift compensation using classifier ensembles},
  author={Vergara, Alexander and Vembu, Shankar and Ayhan, Tuba and Ryan, Margaret A and Homer, Margie L and Huerta, Ram{\'o}n},
  journal={Sensors and Actuators B: Chemical},
  volume={166},
  pages={320--329},
  year={2012},
  publisher={Elsevier}
}

@ARTICLE{VirtanenGommersOliphant2020,
  author  = {Virtanen, Pauli and Gommers, Ralf and Oliphant, Travis E. and
            Haberland, Matt and Reddy, Tyler and Cournapeau, David and
            Burovski, Evgeni and Peterson, Pearu and Weckesser, Warren and
            Bright, Jonathan and {van der Walt}, St{\'e}fan J. and
            Brett, Matthew and Wilson, Joshua and Millman, K. Jarrod and
            Mayorov, Nikolay and Nelson, Andrew R. J. and Jones, Eric and
            Kern, Robert and Larson, Eric and Carey, C J and
            Polat, {\.I}lhan and Feng, Yu and Moore, Eric W. and
            {VanderPlas}, Jake and Laxalde, Denis and Perktold, Josef and
            Cimrman, Robert and Henriksen, Ian and Quintero, E. A. and
            Harris, Charles R. and Archibald, Anne M. and
            Ribeiro, Ant{\^o}nio H. and Pedregosa, Fabian and
            {van Mulbregt}, Paul and {SciPy 1.0 Contributors}},
  title   = {{{SciPy} 1.0: Fundamental Algorithms for Scientific
            Computing in Python}},
  journal = {Nature Methods},
  year    = {2020},
  volume  = {17},
  pages   = {261--272}
}

@article{WidmerKubat1996,
  title={Learning in the presence of concept drift and hidden contexts},
  author={Widmer, Gerhard and Kubat, Miroslav},
  journal={Machine learning},
  volume={23},
  number={1},
  pages={69--101},
  year={1996},
  publisher={Springer}
}

@article{Zhu2010,
  title={Stream data mining repository},
  author={Zhu, Xingquan},
  journal={URL:http://www.cse.fau.edu/xqzhu/stream.html},
  year={2010}
}

@article{Zliobaite2011,
  title={Combining similarity in time and space for training set formation under concept drift},
  author={{\v{Z}}liobait{\.e}, Indr{\.e}},
  journal={Intelligent Data Analysis},
  volume={15},
  number={4},
  pages={589--611},
  year={2011},
  publisher={SAGE Publications Sage UK: London, England}
}

@inproceedings{Zliobaite2010,
  author = {Zliobaitė, Indrė},
  booktitle = {IEEE int. conf. on Data Mining Workshops (ICDMW 2010)},
  pages = {843 – 850},
  title = {Change with Delayed Labeling: When is it Detectable?},
  year = {2010}
}
